\documentclass[]{interact}

\usepackage{epstopdf}
\usepackage[caption=false]{subfig}
\usepackage{xcolor}

\usepackage[sort]{natbib}
\bibpunct[, ]{(}{)}{;}{a}{,}{,}

\usepackage{lineno}
\usepackage{amsmath}

\usepackage{algorithm}
\usepackage{algpseudocode}
\usepackage{array}
\usepackage{graphicx}
\usepackage{url}
\usepackage{pdflscape}
\usepackage{multirow}
\usepackage{bbm}
\usepackage{lineno}
\linenumbers

\theoremstyle{plain}

\theoremstyle{definition}

\theoremstyle{remark}

\usepackage{geometry}
\begin{document}
\nolinenumbers

\articletype{Research Article}

\title{CATS: Conditional Adversarial Trajectory Synthesis for Privacy-Preserving Trajectory Data Publication Using Deep Learning Approaches}

\author{
\name{Jinmeng Rao\textsuperscript{a}, Song Gao\textsuperscript{a*}\thanks{*A preprint draft and the final version will be available on the International Journal of Geographical Information Science; Corresponding Author: Song Gao. Email: song.gao@wisc.edu}, Sijia Zhu\textsuperscript{a,b}}
\affil{\textsuperscript{a} GeoDS Lab, Department of Geography, University of Wisconsin-Madison, WI\\ \textsuperscript{b} Data Science Institute, Columbia University, New York City, NY }}

\maketitle

\begin{abstract}
The prevalence of ubiquitous location-aware devices and mobile Internet enables us to collect massive individual-level trajectory dataset from users. Such trajectory big data bring new opportunities to human mobility research but also raise public concerns with regard to location privacy. In this work, we present the Conditional Adversarial Trajectory Synthesis (CATS), a deep-learning-based GeoAI methodological framework for privacy-preserving trajectory data generation and publication. CATS applies K-anonymity to the underlying spatiotemporal distributions of human movements, which provides a distributional-level strong privacy guarantee. By leveraging conditional adversarial training on K-anonymized human mobility matrices, trajectory global context learning using the attention-based mechanism, and recurrent bipartite graph matching of adjacent trajectory points, CATS is able to reconstruct trajectory topology from conditionally sampled locations and generate high-quality individual-level synthetic trajectory data, which can serve as supplements or alternatives to raw data for privacy-preserving trajectory data publication. The experiment results on over 90k GPS trajectories show that our method has a better performance in privacy preservation, spatiotemporal characteristic preservation, and downstream utility compared with baseline methods, which brings new insights into privacy-preserving human mobility research using generative AI techniques and explores data ethics issues in GIScience.
\end{abstract}


\section{Introduction}

The prevalence of location-based devices and mobile Internet make it possible to build a massive collection of individual-level trajectory data for human mobility research and various applications in navigation, business intelligence, and public health \citep{gao2015spatio, pappalardo2023future,oliver2020mobile}, which also supports the advances in the integrated science of movement \citep{miller2005measurement,demvsar2021establishing,miller2019towards}. Such trajectory data can be collected in various ways such as mobile phones, GPS devices, social media geo-tags, and public transportation records \citep{yue2014zooming,liu2015social,dodge2020progress}. The large-scale individual trajectory data has been used to portray human mobility across space and time, which brings new opportunities for understanding mobility patterns and laying the basis for many research topics in GIScience and beyond such as modeling human dynamics \citep{sun2016identifying,huang2018location,shaw2020understanding,li2022fine}, traffic analysis and public transport planning \citep{caceres2012traffic,liu2020understanding}, understanding the COVID-19 pandemic impacts on human mobility changes \citep{huang2020twitter,noi2022assessing,long2022associations,chow2021geographic}, spatiotemporal analysis of individual air pollution exposure \citep{li2019dynamic,park2017individual}, tourism management \citep{xu2022understanding}, and natural disaster responses \citep{han2019cities}. Although promising, individual trajectory data also raise public concerns about privacy issues \citep{armstrong2005geographic,mckenzie2019geospatial,kamel2022reconciling}. Improper use of trajectory data may not only violate people's rights to prevent the disclosure of identity and associated sensitive locations, but also bring challenges in behavioral, social, ethical, legal and policy implications \citep{kwan2004protection,kessler2018geoprivacy,gao2019exploring,sieber2016geoai}. Recent works have demonstrated the technical capabilities for the development of a privacy-preserving location-sharing platform \citep{mckenzie2022privyto}.

With the growing public awareness of privacy issues, especially the privacy risks in mobile phone data which involve locations~\citep{de2018privacy}, multiple methods have been developed to address individual trajectory privacy protection issues to prevent users' identity from being identified or their sensitive locations from being revealed. \cite{fiore2020privacy} proposed a taxonomy of privacy principles for trajectory privacy research, namely \textit{Mitigation} (i.e., reducing privacy risks in a database using circumstantial methods, such as identifier removal or obfuscation), \textit{Indistinguishability} (i.e., creating an anonymized database where each record is indistinguishable from the others, such as k-anonymization), and 
\textit{Uninformativeness} (i.e., the difference between the knowledge of the adversary before and after accessing a database must be small, such as differential privacy). Common trajectory privacy protection studies follow one or more of these principles and can be further summarized into four research streams: \textit{De-identification, Geomasking, Trajectory K-anonymization, and Differential Privacy.} 

1)\textit{De-identification} simply removes existing identifiers from trajectory dataset such as users' name. This common practice ensures the de-identified trajectory dataset does not contain any identifiers. However, the privacy risk still exists as other trajectory attributes such as location, time, and socioeconomic attributes might serve as "quasi-identifiers" and still cause users to be re-identified \citep{Chow2011}. 

2)Another group of methods named \textit{Geomasking} changes the locations of trajectory data by shifting, blurring, or masking so that original locations are hidden in uncertainty while some spatial characteristics or relationships still remain in the geomasked locations \citep{armstrong1999geographically, hampton2010mapping, gao2019exploring}. A simple and common practice of geomasking is aggregation: it aggregates trajectory locations into grids, geographic or administrative units so that the original locations are concealed. The main drawback of aggregation, however, is that it is hard to control the trade-off between its privacy guarantee and spatial resolution after aggregation. It was proved that, without largely degrading spatial resolution, it could only provide a weak privacy guarantee and cannot prevent important information of trajectory data from being recovered \citep{de2013unique, xu2017trajectory, gao2019exploring}. Other geomasking methods include Random Perturbation \citep{kwan2004protection, seidl2016privacy}, which adds random noise following a uniform distribution (parameterized by a radius threshold) to the locations in the dataset; Donut Geomasking \citep{hampton2010mapping}, similar to random perturbation but has two radius thresholds to ensure the perturbed locations are within the defined donut area; Gaussian Geomasking \citep{armstrong1999geographically}, which adds random noise following a Gaussian distribution (parameterized by a mean value and a standard deviation) to the locations in the dataset. Recent studies on social media geo-tagged data indicated that, under similar settings, Gaussian geomasking had slightly higher privacy protection effectiveness than random perturbation and aggregation due to its theoretically unbounded noise range \citep{gao2019exploring, rao2020lstm}. 

3)\textit{Trajectory K-anonymization} formulates trajectory privacy protection as a K-anonymity problem and mixes K trajectories or location points in aggregated ways to ensure they are indistinguishable from each other \citep{niu2014achieving,jain2016big,zhu2019private}. A classic method of Trajectory K-anonymization is spatial cloaking. It groups trajectory points from different users into K-anonymized cloaking regions, providing the K-anonymity privacy guarantee \citep{gruteser2003anonymous}. Other methods also utilize a similar logic to achieve K-anonymity by mixing trajectory points within K-anonymized regions \citep{nergiz2008towards, palanisamy2011mobimix}. Based on the idea of K-anonymity, several extensions are also proposed, such as l-diversity \citep{machanavajjhala2007diversity} and t-closeness \citep{li2007t}. 

4)\textit{Differential Privacy (DP)}, proposed by \cite{dwork2006differentialprivacy} as a rigorous mathematical definition of privacy, has been widely used in privacy-preserving data publishing. By adding noise (e.g., Laplace noise) to the query result related to a database, others cannot determine whether a record exists in the database or not. Recently, many studies introduced differential privacy into trajectory privacy protection. For example,  \cite{andres2013geo} proposed differential-privacy-based Geo-indistinguishability for trajectory data publishing. By adding controlled random noise, it can prevent the real personal locations of users from being identified. \cite{li2017achieving} proposed a differentially private trajectory data publishing algorithm with a bounded Laplace noise generation algorithm and a trajectory merging algorithm. \cite{deldar2018pldp} introduced a concept of personalized-location differential privacy (PLDP) for trajectory databases, and devised a differentially private algorithm to provide non-uniform privacy guarantees. \cite{chen2019real} introduced a differential-privacy-based trajectory privacy protection framework with real-path reporting in mobile crowdsensing, which was able to preserve trajectory privacy under Bayesian inference attacks.

While in certain ways the aforementioned approaches can be applied to protect trajectory privacy, a major challenge remains, which is the level of trajectory privacy guarantee and  the trade-off between privacy and trajectory data utility. Despite the diversity, most of existing approaches try to add uncertainty (i.e., noise) to the locations of trajectories. However, the underlying raw data distribution still exists and so does privacy risks. Under certain situations, an attacker is still able to uncover the added uncertainty and reveal the raw data distribution. As \cite{andres2013geo} pointed out, while adding controlled noise to a user's location might achieve geo-indistinguishability on a single location, the privacy guarantee may be downgraded when applying the method to a group of locations as they might be correlated. Similarly, if attackers possess sufficient auxiliary information, they may be able to eliminate the added noise in trajectories and reveal the real data. While extreme, these failure cases suggest that privacy risks persist if the raw data distribution is not adequately protected. Current location-level or trajectory-level privacy protection methods do not address this issue.
Thus, the main research question (RQ) that we want to investigate in this work is: \textit{Given a trajectory dataset with certain spatiotemporal distribution, can we provide a better trajectory privacy guarantee using synthetic data while balancing the trade-off between trajectory privacy, spatiotemporal characteristics preservation, and data utility?}

In this work, empowered by Geospatial Artificial Intelligence (GeoAI) such as spatially explicit machine learning and deep learning approaches~\citep{janowicz2019geoai,mai2022symbolic}, we present the CATS, a deep-learning-based Conditional Adversarial Trajectory Synthesis framework for privacy-preserving trajectory generation and publication. By leveraging deep learning approaches such as conditional adversarial neural network training, CATS is able to generate high-quality individual-level synthetic trajectory data from k-anonymized and aggregated human mobility matrices, which can serve as supplements and alternatives to raw data for privacy-preserving data publication. The trajectory generation task, which aims to generate individual synthetic trajectories with realistic mobility patterns, is one of the well-defined mobility tasks along with origin-destination flow generation, crowd flow prediction, and next-location prediction~\citep{luca2021survey}. Compared with traditional mechanistic generative models~\citep{song2010modelling, jiang2016timegeo, pappalardo2018data}, deep-learning approaches can capture complex and non-linear relationships in the data, thus generating more realistic trajectories. More importantly, limited attention has been paid to the privacy protection capability in existing deep learning approaches. Although recent works have applied deep neural networks (e.g., generative adversarial network, GAN) in synthetic trajectory data generation such as LSTM-TrajGAN~\citep{rao2020lstm}, TrajGAIL~\citep{choi2021trajgail}, TrajGen~\citep{cao2021generating}, Deep Gravity~\citep{simini2021deep}, and MoGAN~\citep{giovanni2022generating}, the key differences between this work and existing ones are as follows: 1) we provide a distributional-level K-anonymity privacy guarantee and verify the privacy protection effectiveness by experiments, which was not investigated in TrajGAIL, TrajGen, Deep Gravity, or MoGAN. LSTM-TrajGAN also examines its privacy protection effectiveness by experiments, but it did not consider the conditional spatiotemporal data distribution; 2) we achieve conditional trajectory generation by using aggregated human mobility distribution as a condition and model trajectory global context using the attention-based mechanism, which was not supported in other works; and 3) we generate individual-level synthetic trajectory data to preserve spatiotemporal mobility patterns of raw trajectory data at a city scale. TrajGAIL generate synthetic data to simulate vehicle driving behaviors only in a regional road network. TrajGen and MoGAN map mobility trajectories or aggregated flows into a single image, where time information is lost and needs to be predicted or recovered separately.

The remainder of the paper is organized as follows. Section 2 introduces our methodology, including the overall design of the CATS framework, component details, and the optimization objectives. Section 3 introduces the experiments and results where we conduct comprehensive tests from the perspectives of trajectory privacy preservation, spatiotemporal characteristics preservation, and downstream utility. Section 4 discusses the factors affecting the framework performance, privacy guarantee, and some limitations of this work. Section 5 outlines the conclusions and our future works. 

\section{Methodology}

In this section, we introduce the Conditional Adversarial Trajectory Synthesis (CATS), a deep-learning-based GeoAI framework that supports privacy-preserving synthetic trajectory data generation based on aggregated human mobility distributions and generative AI techniques. We first explain several preliminary concepts and definitions involved in this work, and then illustrate the methodological framework in detail, including model design and optimization objectives to guarantee privacy protection and trajectory data utility.

\subsection{Preliminary}

\paragraph*{Trajectory:} In geographic information science, a trajectory $\mathcal{T}$ is considered as a sequence of ordered points that describe a discrete trace of a moving object in geographical space during a specific time period. A point in a trajectory can be defined as a tuple $(l,t)$, where $l$ and $t$ denote the location and timestamp when the point was sampled. From a geographic perspective, a location $l$ can be described by a tuple $(x, y)$, where $x$ and $y$ denote the geographic coordinates of the point (i.e., latitude and longitude). A trajectory $\mathcal{T}$ therefore can be represented as $\mathcal{T}=\{(l_i, t_i)\}$, showing how a person or an animal moves from one location to another sequentially. A trajectory dataset is denoted as $\mathcal{X}=\{\mathcal{T}_{i}\}$.

\paragraph*{Spatiotemporal Mobility Matrix:}A spatiotemporal mobility matrix $\mathcal{M}^{stm}$ defined in this research is a three-dimensional matrix describing human mobility's discrete spatiotemporal probability distributions in a 3D cube (x, y, t), where $x$ and $y$ dimensions indicate geographic coordinates and its $t$ dimension indicates time. Each slice of $\mathcal{M}^{stm}$ along the time dimension (e.g., at time $t_j$) is a two-dimensional probability distribution (denoted as $\mathcal{P}_{t_j}$), revealing the spatial distribution of a person's historical locations at that particular time in a day.

\begin{figure}[h]
  \centering
  \includegraphics[width=0.9\linewidth]{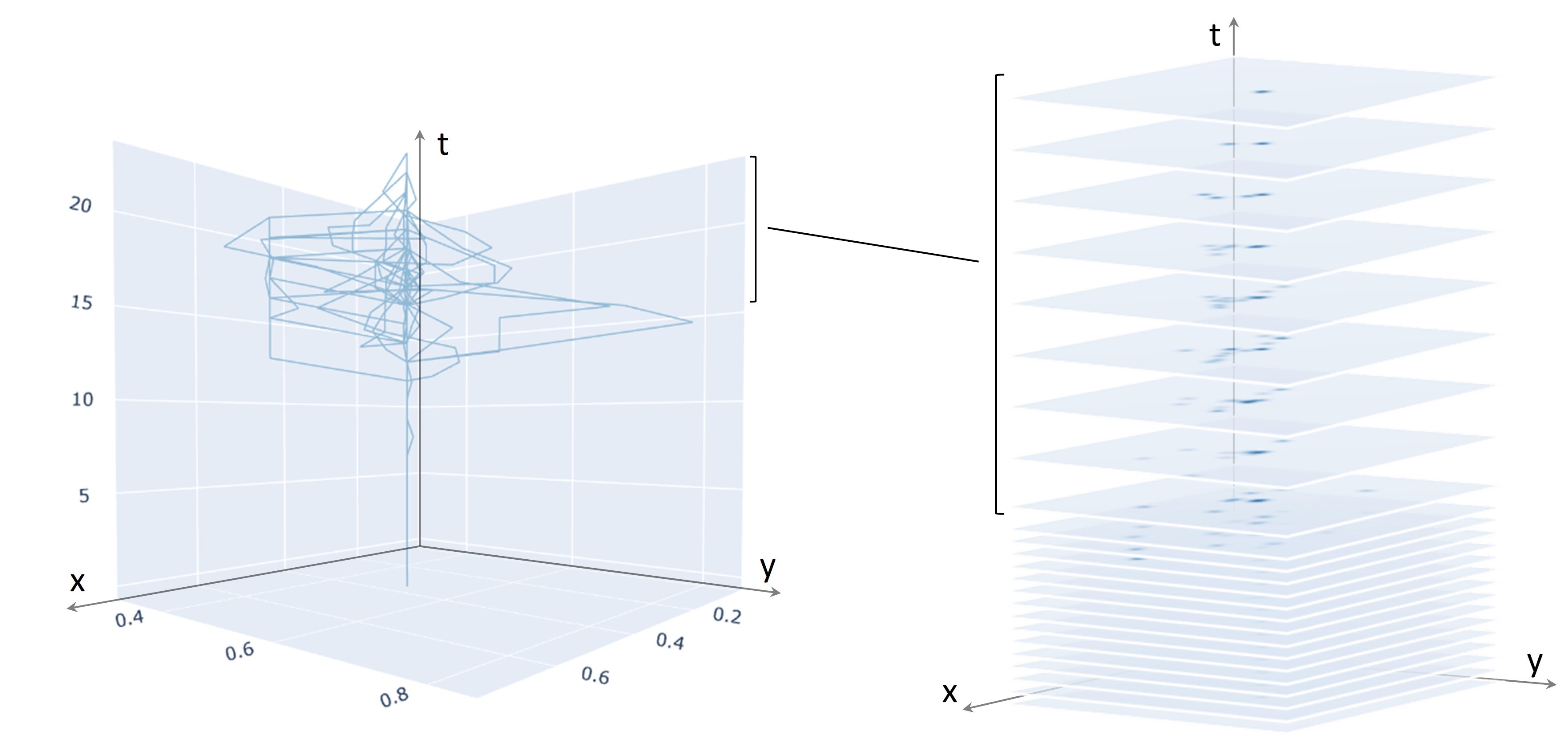}
  \caption{A visual example of a continuous three-dimensional space recording daily trajectories of a person (left) and the discretized matrix describing corresponding human mobility's probability distributions in a 3D cube.}
  \label{fig:visual_example}
\end{figure}

Figure \ref{fig:visual_example} shows an example of using a continuous three-dimensional space to record the daily trajectories of a person during a certain time range. In this case, the $x$ and $y$ axis represent two geographic directions. The $t$ axis indicates the time in a day. Given a spatial resolution (e.g., 500 m) and a time resolution (e.g., 1 hour), the space can be discretized into a three-dimensional matrix, where the value of each cell indicates the number of an individual's trajectory points that the cell contains at that time period. After normalizing all the slices along the time axis, we obtain the spatiotemporal mobility matrix of that person, where each time slice is a two-dimensional probability distribution of the person's historical locations at that time period. By constructing the spatiotemporal mobility matrix of a person, we can learn the person's mobility patterns, such as activity zone, location preference, etc., and further infer privacy-sensitive locations of the person, such as their home location.

\subsection{Conditional Adversarial Trajectory Synthesis}

\begin{figure}[h]
  \centering
  \includegraphics[width=\linewidth]{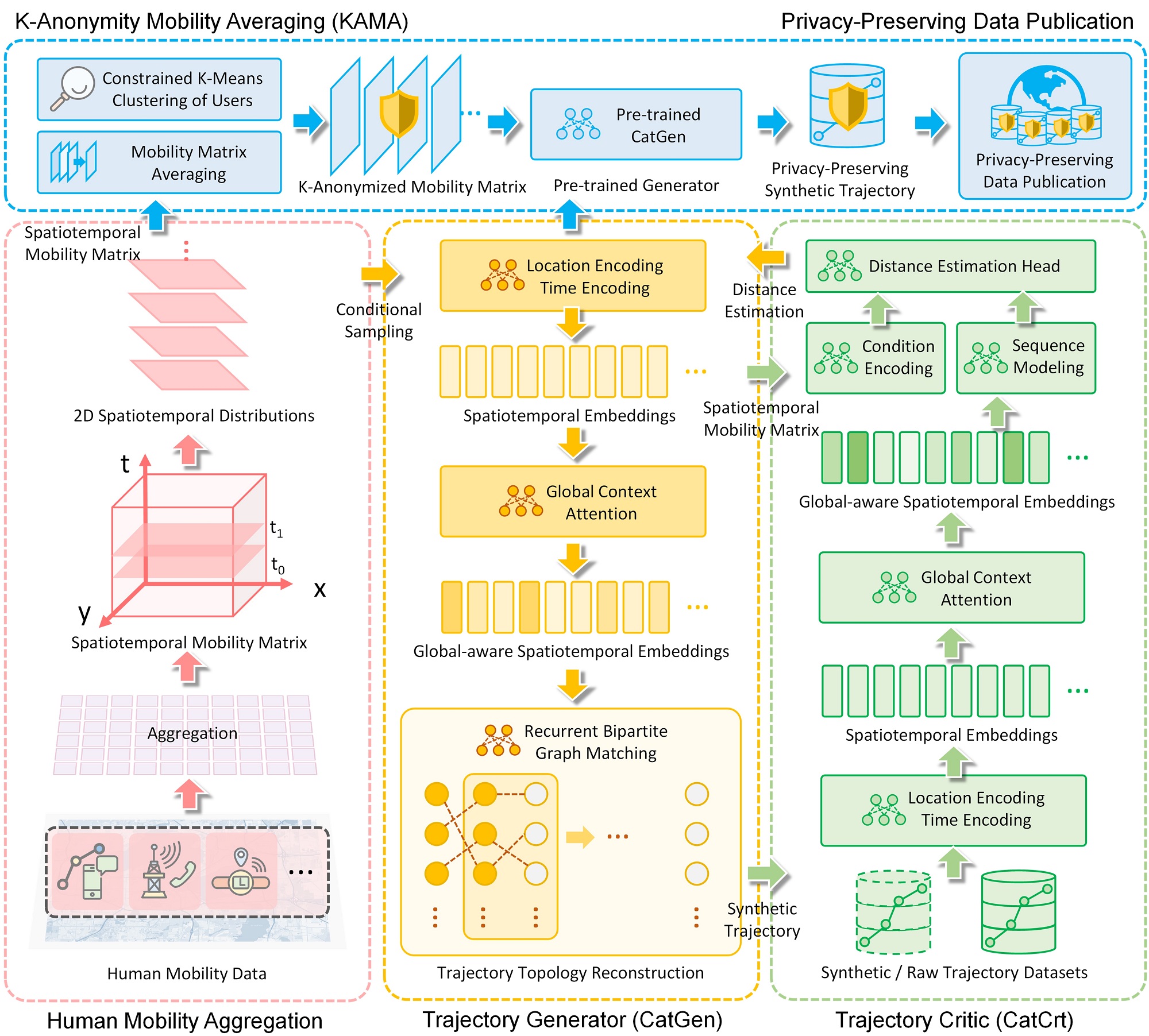}
  \caption{The overall design of the Conditional Adversarial Trajectory Synthesis (CATS) framework.}
  \label{fig:overall_temp}
\end{figure}

Given a person's spatiotemporal mobility matrix $\mathcal{M}^{stm}_u$, the goal of CATS is to generate privacy-preserving synthetic individual trajectory data with realistic mobility patterns, including spatiotemporal distribution and topology (i.e., sequence of locations). The overall design of the CATS framework is depicted in Figure \ref{fig:overall_temp}. There are four main parts in the framework, namely Human Mobility Aggregation, Trajectory Generator (CatGen), Trajectory Critic (CatCrt), and K-Anonymity Mobility Averaging (KAMA):

\paragraph*{Human Mobility Aggregation:} the purpose of this part is to aggregate individual human mobility data (e.g., individual-level daily GPS trajectory data from mobile phones and wearable devices) into spatiotemporal mobility matrices. Specifically, we encode individual-level daily trajectory data of each person into a grid system with specific spatial and temporal resolutions, which generates a discrete spatiotemporal mobility matrix $\mathcal{M}^{stm}_u$ of a person $u$ in a three-dimensional space. The $x$ and $y$ dimensions represent two geographic directions, while the $t$ dimension represents time. Note that each time slice of $\mathcal{M}^{stm}_u$ represents a two-dimensional probability distribution that reveals the spatial distribution of the historical locations of the person $u$ at that particular time of day.

\paragraph*{Trajectory Generator (CatGen):} given $\mathcal{M}^{stm}_u$, the purpose of CatGen is to generate synthetic trajectories with realistic mobility patterns accordingly. CatGen first samples trajectory points at each time conditional on $\mathcal{M}^{stm}_u$, and then it utilizes deep learning techniques including encoding layers, self-attention mechanism, and recurrent neural networks to learn how to reconstruct synthetic trajectories with realistic mobility patterns.

\paragraph*{Trajectory Critic (CatCrt):} given $\mathcal{M}^{stm}_u$, the purpose of CatCrt is to estimate the difference between the probability distribution of raw trajectory data and synthetic trajectory data generated by CatGen conditional on their corresponding $\mathcal{M}^{stm}_u$. The feedback from CatCrt is passed to CatGen to help it learn how to create better mobility patterns, which leads to an adversarial training process (i.e., an intelligent two-player minimax game).

\paragraph*{K-Anonymity Mobility Averaging (KAMA):} the purpose of KAMA is to obtain the K-anonymized spatiotemporal mobility matrices from raw trajectory data, which lays the basis for privacy-preserving trajectory generation and publication. KAMA first cluster people into smaller groups with a miminum size of $K$ people. Then KAMA performs mobility matrix averaging among each group to obtain the K-anonymized spatiotemporal mobility matrices for each group. By leveraging a pre-trained CatGen, CATS is able to generate realist synthetic trajectory data with K-anonymity from such K-anonymized matrices, which could serve as supplemental or alternative data for publication on premise of trajectory privacy preservation.

\subsubsection{CatGen: Trajectory Generator} \label{section:CatGen}

As shown in Figure \ref{fig:overall_temp}, CatGen has four layers, namely a Conditional Sampling layer, a Spatiotemporal Encoding layer, a Global Context Attention layer, and a Recurrent Bipartite Graph Matching layer. We explain these layers below in detail.

\paragraph*{Conditional Sampling:} In this layer, given a spatiotemporal mobility matrix $\mathcal{M}^{stm}_u$ aggregated from the historical trajectory data of a user $u$ with a spatial resolution of $N \times N$ and a temporal size of $t > 0$, we sample a fixed-size set of locations $\mathcal{S}_{t_j}$ conditional on the two-dimensional probability distribution slice $\mathcal{P}_{t_j}$ in $\mathcal{M}^{stm}_u$ at time $t_j$:

\begin{equation}
 \mathcal{S}_{t_j} = \{l_i | l_i \leftarrow \mathcal{P}_{t_j}\}, \mathcal{P}_{t_j} \in \mathcal{M}^{stm}_u
\end{equation}

where $l_i$ is the $i$-th sampled location conditional on $\mathcal{P}_{t_j}$. Specifically, for each $l_i$ there exists a corresponding probability value $p_{l_i}\in \mathcal{P}_{t_j}$ that denotes the likelihood of $l_i$ being selected during the sampling process. Such a conditional sampling process yields a time-ordered sequence of trajectory location sets $\{\mathcal{S}_{t_j}\}$, and the location frequency counted in $\mathcal{S}_{t_j}$ may naturally reflect the spatial distribution shown in $\mathcal{P}_{t_j}$. This ensures that the spatial distribution of the locations in the synthetic trajectories at each time is very close, if not identical, to that of the corresponding raw trajectories.

\paragraph*{Spatiotemporal Encoding:} In this layer, given the sampled time-series of trajectory location sets $\{\mathcal{S}_{t_j}\}$, we perform location encoding $Enc_{l}$ and time encoding $Enc_{t}$ on the spatial dimension and temporal dimension of $\{\mathcal{S}_{t_j}\}$ respectively to get the location embeddings and time embeddings, which are learning-friendly for downstream deep learning models \citep{rao2020lstm,mai2022review}. Specifically, for both $Enc_{l}$ and $Enc_{t}$, we use a fully connected neural network layer (e.g., a dense layer) followed by a Rectified Linear Unit (ReLU) that adds non-linearity. Note that the weights of encoding layers are shared across location and time inputs to ensure the encoding consistency. We further concatenate the output embeddings together as a spatiotemporal embedding.

\begin{equation}
Embed_{l_i, t_j} = concat(Enc_{l}(l_i), Enc_{t}(t_j))
\end{equation}

where $concat(\cdot)$ is a concatenation operator that concatenates two matrices into one; $Embed_{l_i, t_j}$ is the concatenated spatiotemporal embedding that carries both spatial and temporal information of a trajectory point. Furthermore, the spatiotemporal embeddings of all the sampled trajectory points at time $t_j$ can be denoted as $Embed_{t_j}$.

\paragraph*{Global Context Attention:} Each location $l_i$ in $\mathcal{S}_{t_j}$ is separately sampled and encoded without any interaction with other sampled locations, leading to the absence of a sense of awareness among them. We define the global-scale relationships among all trajectory points at a specific time step as the Global Context, which we believe is important in the matching-based trajectory topology reconstruction process. By capturing such global context at each time step and integrating it into spatiotemporal embeddings, we can enhance the process of reconstructing realistic synthetic trajectories. In this layer, we apply the Multi-Head Self-Attention (MHSA) \citep{vaswani2017attention} on $Embed_{t_j}$ at each time step to capture the global context among trajectory points. MHSA is a type of self-attention mechanism that computes an attention-weighted representation capturing the intra attention inside given inputs. We choose MHSA to capture the global context mainly because: 1) the attention mechanism allows each member (e.g., spatiotemporal embedding in our case) in the input to interact with the others and aggregates the interactions as attention scores, which aligns with our purpose to capture each trajectory location's global context; and 2) existing research suggests that MHSA, as a widely used attention component in attention-based models, enables the model to jointly attend to information from different representation subspaces at different positions, thereby providing rich representations and enhancing training stability~\citep{liu2021multi}. One attention function introduced by \cite{vaswani2017attention} is the Scaled Dot-Product Attention:

\begin{equation}
Attention(Q, K, V) = softmax(\frac{QK^T}{\sqrt{d_k}})V
\end{equation}

where $Q$, $K$, and $V$ are a matrix of queries, keys, and values, which serve as different linear projections of the input spatiotemporal embeddings; $\sqrt{d_k}$ is the dimension of both keys and queries. MHSA obtains multiple different representations of $Q$, $K$, and $V$ via linear projection and performs the Scaled Dot-Product Attention for each representation (i.e., multiple heads in parallel), respectively:

\begin{equation}
\begin{aligned}
MHSA^{h}(Q, K, V) = concat(head_1, ..., head_h)W^O\\
head_i = Attention(Q W^Q_i, K W^K_i, V W^V_i)
\end{aligned}
\end{equation}

where $W^Q_i$, $W^K_i$, and $W^V_i$ are the linear projection parameters for $head_i$; $h$ denotes the number of heads in MHSA; $W^O$ is the linear projection parameter for MHSA; $concat(\cdot)$ is the concatenation operator that concatenates multiple matrices into one. By leveraging MHSA, we are able to capture the trajectory global context information and integrate it into the spatiotemporal embeddings:

\begin{equation}
\begin{aligned}
Embed^{gc}_{t_j} = LayerNorm(Embed_{t_j} + MHSA^{h}(Embed_{t_j}, Embed_{t_j}, Embed_{t_j}))
\end{aligned}
\end{equation}

where $Embed_{t_j}$ denotes spatiotemporal embeddings of all the sampled trajectory points at time $t_j$; $MHSA^{h}(Embed_{t_j}, Embed_{t_j}, Embed_{t_j})$ captures global context from $Embed_{t_j}$; $Embed^{gc}_{t_j}$ denotes spatiotemporal embeddings at time $t_j$ enhanced by their global context (i.e., global-aware spatiotemporal embeddings). Following a similar encoder design in \cite{vaswani2017attention}, We apply the Layer Normalization to normalize the output to improve training stability.

\paragraph*{Recurrent Bipartite Graph Matching:} 
In this layer, we sequentially assemble sampled trajectory points across time into individual synthetic trajectories (i.e., trajectory topology reconstruction) via embedding-based matching. This recurrent matching layer is necessary because the global-aware spatiotemporal embeddings at different time steps $Embed^{gc}_{t_j}$ are generated separately, and therefore the relationship between every two temporally adjacent sets of spatiotemporal embeddings is not preserved. In other words, the realistic trajectory location topology is not preserved in $\{Embed^{gc}_{t_j}\}$. Hence, to generate synthetic trajectory data with realistic movement patterns, we need to reconstruct the location topology of trajectories by re-establishing realistic matching relationship between every two temporally adjacent sets of spatiotemporal embeddings. We formulate the matching between two embedding sets as a \textit{Bipartite Graph Matching} problem: given a complete Bipartite Graph denoted as $G=(U,V;E)$, where $U$ and $V$ denote two disjoint and independent vertex parts and $E$ denotes the edges between $U$ and $V$. Each edge $e(u,v) \in E$ has a non-negative cost $c(u, v)$, where $u \in U$, $v \in V$. The goal is to find a \textit{perfect matching} (i.e., each $u \in U$ is matched to one and only one $v \in V$, and vice versa) with a minimized total cost. In our case, given two temporally adjacent spatiotemporal embedding sets $Embed^{gc}_{t_j}$ and $Embed^{gc}_{t_j+1}$, each spatiotemporal embedding is a vertex in the bipartite graph, and thus $U_{t_j}=Embed^{gc}_{t_j}$ and $U_{t_j+1}=Embed^{gc}_{t_{j+1}}$. Note that it is a balanced bipartite graph since $|U_{t_j}|=|U_{t_j+1}|$. The cost in our case denotes the distance between two embedding vectors, and we use Euclidean distance as the cost function. Specifically, we directly calculate L2 norm between two embedding vectors. Since the matching between trajectory points at different time steps relies on not only their spatiotemporal embeddings at the current step, but also the historical matched embeddings from previous steps (e.g., historical mobility of users), we further incorporate a Recurrent Neural Network (RNN) to model the movement history (i.e., embeddings of past trajectory points) reconstructed till time $t_j$ and update the vertices in $U_{t_j}$ recurrently. Then we find the perfect matching between $U_{t_j}$ and $U_{t_j+1}$ where the sum of costs is minimized, which achieves \textit{Recurrent Bipartite Graph Matching} and forms a reconstruction plan of trajectory topology. Note that during the training phase, CatGen is expected to acquire the ability to generate suitable spatiotemporal embedding representations in such a way that the matching cost of spatiotemporal embedding pairs, which are from the same trajectory and temporally adjacent, to be lower than that of other pairs. The matching cost depends on both spatial and temporal information of the embeddings. For mobility history modeling, we use the Gated Recurrent Units (GRUs), a common RNN structure based on the gating mechanism with strong time-series sequence modeling ability~\citep{zhao2019t}. The definition of GRUs is as below:

\begin{equation}
\begin{aligned}
z_t &= \sigma_g(W_{z} x_t + Y_{z} h_{t-1} + b_z) \\
r_t &= \sigma_g(W_{r} x_t + Y_{r} h_{t-1} + b_r) \\
\hat{h}_t &= \phi_h(W_{h} x_t + Y_{h} (r_t \odot h_{t-1}) + b_h) \\
h_t &=   z_t \odot \hat{h}_t + (1-z_t) \odot h_{t-1}
\end{aligned}
\end{equation}

where $x_t$ denotes a input vector (e.g., an input spatiotemporal embedding in our case) at each time step; $h_t$ denotes an output vector (e.g., an output spatiotemporal embedding in our case) at each time step; $\hat{h}_t$ denotes a candidate activation vector; $z_t$ denotes an update gate vector; $r_t$ a reset gate vector; $W$, $Y$ and, $b$ denotes parameter matrices and vector; and $\sigma_g$ and $\phi_h$ denote Sigmoid activation function and Tanh function. It is worth noting that the output vector $h_t$ at each time step not only contains the information from the current input vector $x_t$, but also incorporates the information from the previous output vector $h_{t-1}$. Therefore, $h_t$ can be considered as a representation of the input sequence (e.g., an embedding sequence of a trajectory) up to time $t$.

As a common practice, we formulate the problem of finding perfect matching with minimized total cost in a bipartite graph as an assignment problem, particularly a Linear Sum Assignment (LSA) problem~\citep{kuhn1955hungarian}: given two sets $\mathcal{S}_{a}$ and $\mathcal{S}_{b}$, a boolean matrix $X$ describing the assignment between $\mathcal{S}_{a}$ and $\mathcal{S}_{b}$ (e.g., $X_{i,j}=1$ indicates $i$ and $j$ are matched, $i \in \mathcal{S}_{a}$, $j \in \mathcal{S}_{b}$), and a matrix $C$ describing the cost of the assignment between $\mathcal{S}_{a}$ and $\mathcal{S}_{b}$ (i.e., $C_{i,j}$ is the cost of the matching between $i \in \mathcal{S}_{a}$ and $j \in \mathcal{S}_{b}$), we want to find the assignment plan with minimized total cost:

\begin{equation}
\min \sum_i \sum_j C_{i,j} X_{i,j}
\end{equation}

Note that during matching each row should be matched to at most one column, and each column should be matched to at most one row. The LSA problem can be solved efficiently in polynomial time by the Hungarian algorithm~\citep{kuhn1955hungarian}. The Hungarian algorithm is a well-known algorithm in computer science used to find the matching in a bipartite graph with minimized cost. It is a preferred choice due to its polynomial time complexity and availability of mature implementations. We apply the Hungarian algorithm recurrently on every two temporally adjacent embedding sets to get the matching relationship with minimized matching cost, where the cost is calculated by Euclidean distance between embedding vectors. After each matching step, we sort the matched embedding sets based on the matching relationship to ensure the they are element-wise aligned such that the inputs GRUs take are in the same order for sequential modeling. Following the completion of the entire recurrent matching process, a reconstruction plan is established to assemble synthetic trajectories. Based on the plan, the final synthetic trajectories are then assembled from sampled trajectory points. Please note that the synthetic trajectories maintain the same form as the raw trajectories $\mathcal{T}=\{(l_i, t_i)\}$ rather than just embeddings.

After going through the above-mentioned four layers in CatGen, a user's spatiotemporal mobility matrix $\mathcal{M}^{stm}_u$ would be dis-aggregated spatially and temporally into a set of individual-level synthetic trajectories (controlled by sample size). Because of the conditional sampling nature, the spatiotemporal distribution of raw trajectory points at each time step is preserved in synthetic trajectory points. The reconstruction of trajectory topology, however, needs to be further learned and optimized with the help of the Trajectory Critic (CatCrt).

\subsubsection{CatCrt: Trajectory Critic}

CatCrt accepts raw or synthetic trajectory data of a user $u$ as inputs and estimates the difference between the probability distributions of the raw and synthetic data, given the user's spatiotemporal mobility matrix $\mathcal{M}^{stm}_u$. The feedback from CatCrt plays a vital role in the optimization of CatGen in section \ref{section:CatGen}, as it informs CatGen about the degree of discrepancy between the synthetic and raw data distributions. As shown in Figure \ref{fig:overall_temp}, CatCrt has five layers, namely a Spatiotemporal Embedding layer, a Condition Embedding layer, a Global Context Attention layer, a Recurrent Sequence Modeling layer, and a Distance Estimation Head layer. Overall the CatCrt shares similar technical architecture with CatGen but differs in its functionality.

\paragraph*{Spatiotemporal Encoding:} Same as CatGen, location encoding $Enc_{l}$ and time encoding $Enc_{t}$ on the spatial dimension and temporal dimension of $\{\mathcal{S}_{t_j}\}$ to get the location embeddings and time embeddings, and we further concatenate them together to get the spatiotemporal embeddings $Embed_{l_i, t_j}$.

\paragraph*{Condition Encoding:} Note that CatCrt needs to estimate the distance between the raw data distribution and synthetic data distribution conditional on spatiotemporal mobility matrix $\mathcal{M}^{stm}_u$, we further embed $\mathcal{M}^{stm}_u$ into a condition embedding $Embed^{cond}_{u}$ and use it for distance estimation. Since $\mathcal{M}^{stm}_u$ can be regarded as an image with a pixel size of $N \times N$ and $t$ channels, an intuitive idea is to extract features from it using Convolutional Neural Network which can efficiently capture image features. In practice, we use a Residual Convolutional Neural Network (ResNet) \citep{he2016deep} to transform $\mathcal{M}^{stm}_u$ into $Embed^{cond}_{u}$. ResNet is a widely-used convolutional neural network architecture due to its efficiency and has been applied as an encoding backbone to extract features in various models (e.g., autoencoder, GAN) in computer vision tasks (e.g., image recognition, object detection). Specifically, the spatiotemporal mobility matrices are processed through the convolutional and pooling layers in ResNet, resulting in higher-dimensional embedding vectors with width and height dimensions reduced to 1.

\paragraph*{Global Context Attention:} For the same reasons described in section \ref{section:CatGen}, in this layer we also apply the MHSA on all the spatiotemporal embeddings $Embed_{t_j}$ at each time step to capture the global context information and integrate it into $Embed_{t_j}$, generating global-aware spatiotemporal embeddings $Embed^{gc}_{t_j}$ (i.e., the attention-weighted representations of $Embed_{t_j}$).

\paragraph*{Recurrent Sequence Modeling:} Similar to CatGen, we need to model the movement history captured in trajectories in CatCrt for distance estimation. Thus, in this layer we also add the GRUs for recurrent modeling of a time-ordered sequence of $\{Embed^{gc}_{t_j}\}$. The sequence modeling of CatCrt differs from that of CatGen as here we adopt a many-to-one structure where we only keep the last output from GRU for the following distance estimation layer.

\paragraph*{Distance Estimation Head:} In this layer we estimate the distance between the raw trajectory data distribution and the synthetic data distribution based on the output of the Recurrent Sequence Modeling layer and the corresponding condition embedding $Embed^{cond}_{u}$. We first concatenate the sequence modeling outputs $Embed^{seq}_{u}$ and $Embed^{cond}_{u}$:

\begin{equation}
Embed^{pred}_{u} = concat(Embed^{seq}_{u}, Embed^{cond}_{u})
\end{equation}

where $Embed^{pred}_{u}$ is the embedding concatenated from $Embed^{seq}_{u}$ and $Embed^{cond}_{u}$; $concat(\cdot)$ is the concatenation operator that concatenates two matrices into one. We then feed $Embed^{pred}_{u}$ into the distance estimation head, which consists of two fully connected layers with a ReLU in the middle. Since CatCrt is designed to complete a regression task that is unbounded (i.e., fitting the distance between two distributions), we do not add a sigmoid function at the end of this component.

\subsubsection{KAMA: K-Anonymity Mobility Averaging}

The CatGen and CatCrt components are designed to ensure that the synthetic trajectory data could preserve mobility patterns of raw trajectory data to guarantee spatiotemporal characteristics and data utility. K-Anonymity Mobility Averaging (KAMA) is another key component in CATS that ensures the generated synthetic trajectory data provide K-anonymity, therefore preserving trajectory privacy. KAMA consists of two steps, namely Constrained Clustering of Users and Mobility Averaging.

\subsubsection{Constrained Clustering of Users}

As we aim to conduct K-anonymization on individuals' mobility matrices, it is essential to ensure that each person can be assigned to a user cluster comprising at least $K$ individuals. Thus, we perform the Constrained K-Means Clustering~\citep{bradley2000constrained} on users' centroids of trajectory points to cluster users. We use the users' centroids due to the simplicity and having been widely used in human mobility-based user clustering research~\citep{yuan2014measuring,ashouri2015cloaked,gao2019exploring}. Constrained K-Means clustering is a variant of the standard K-Means clustering algorithm. Compared with other commonly used location clustering algorithms such as DBSCAN~\citep{ester1996density}, it enables the specification of a minimum size for each cluster. Specifically:

\begin{equation}
\begin{aligned}
\min_{C, T} \sum^{N}_{i=1} \sum^{H}_{h=1} T_{i,h} \cdot (\frac{1}{2} ||x^i - C^h||^2_2 )\\
s.t.
\begin{aligned}
\sum^{N}_{i=1} T_{i,h} \ge \tau_h, h=1,...,H,\\
\sum^{H}_{h=1} T_{i,h} = 1, i=1,...,N,\\
T_{i,h} \ge 0, i=1,...,n, h=1,...,H.
\end{aligned}
\end{aligned}
\end{equation}

where $C^h$ is the center of the $h$-th cluster $Cl^h$; $\tau_h$ is the minimum number of points (e.g., user centroids) the $h$-th cluster should contain (e.g., a constrain); $H$ is the number of desired clusters; $N$ is the total number of points; $x^i$ is the $i$-th points; and $T_{i,h}$ is the selection variable, $T_{i,h}=1$ if $x^i$ is closest to $C^h$, otherwise $T_{i,h}=0$. By appropriately configuring $H$ and $\tau_h$, we can effectively group each user with a minimum number of their relatively close neighboring individuals into a cluster. Additionally, clustering users' centroids enables the grouping of individuals with similar activity zones~\citep{liu2021activity}, promoting the preservation of users' mobility patterns following mobility averaging. It is worth noting that this approach differs from acquiring K-Nearest Neighbors (KNN) since the clusters obtained from Constrained K-Means Clustering do not overlap each other (i.e., hard clustering). This property guarantees that by setting $\tau_h=K$ ($h=1,...,H$), each cluster $Cl^h$ comprises at least $K$ users, which facilitates the implementation of K-anonymity. Regarding the selection of $K$, according to the Privacy Technical Assistance Center (PTAC) of the U.S. Department of Education, many statisticians consider a cell size of at least 3 to prevent disclosure, though larger minimums (e.g., 5 or 10) may be used to further mitigate disclosure risk~\citep{daries2014privacy}.

\subsubsection{Mobility Averaging}

After getting the user clusters $\{Cl^h\}$, we average their spatiotemporal mobility matrices $\mathcal{M}^{stm}_u$ and assign the corresponding averaged matrix back to each user inside each cluster. For simplicity, we sum all the mobility matrices by element-wise in each cluster $Cl^h$ and then divide them by the size of clusters $|Cl^h| \geq K$, respectively:

\begin{equation}
\mathcal{M}^{stm}_{avg} = \frac{1}{|Cl^h|}\sum^{|Cl^h|}_{u=1} \mathcal{M}^{stm}_u
\end{equation}

$\mathcal{M}^{stm}_{avg}$ reflects the mobility patterns of all $K$ users in the cluster, providing a K-Anonymity privacy guarantee from the distributional level. Naturally, as a result, feeding ${\mathcal{M}^{stm}_{avg}}$ into a pre-trained CatGen can generate K-anonymized synthetic trajectory data.

\subsubsection{Optimization Objectives} \label{sec:optimization}

The optimization of CATS aims to minimize the difference between the data distribution of raw trajectories and that of the synthetic trajectories, particularly the probability distributions of how their trajectory points are connected (i.e., trajectory topology), by updating deep learning parameters in CatGen and CatCrt through conditional adversarial training. During inference, a well-trained CATS is able to generates synthetic trajectory data that are similar to the raw trajectory data in terms of mobility patterns, given a spatiotemporal mobility matrix of a user. The optimization objective of CATS follows the design of the Wasserstein GAN~\citep{arjovsky2017wasserstein}, which is an more efficient variant of the original GAN and has been proven to improve learning stability and prevent mode collapse (i.e., the generator of GAN only produces a single type of output or a small set of outputs, resulting in all of the generated trajectories very similar or even identical). The original GAN learns to optimize the following objective:

\begin{equation}
\min_{G}\max_{D}\mathbb{E}_{x\sim p_{\text{data}}(x)}[\log{D(x)}] +  \mathbb{E}_{z\sim p_{\text{z}}(z)}[1 - \log{D(G(z))}]
\end{equation}

where $p_{data}(x)$ denotes raw data distribution; $p_{z}(z)$ denotes a prior on noise variables; $D(x)$ denotes the probability that $x$ came from $p_{data}(x)$; $G(z)$ denotes a mapping from $p_{z}(z)$ to $p_{data}(x)$. The generator $G$ aims to minimize $\mathbb{E}_{z\sim p_{z}(z)}[log(1-D(G(z)))]$ while the discriminator $D$ aims to maximize $\mathbb{E}_{x\sim p_{data}(x)}[logD(x)] + \mathbb{E}_{z\sim p_{z}(z)}[log(1-D(G(z)))]$, resulting in a two-player minimax game. It is proved that optimizing the original GAN objective is in fact minimizing the Jensen–Shannon divergence (JSD) between two distributions~\citep{arjovsky2017towards}. One issue of JSD is that it returns a constant value when the two distributions do not align and may result in vanishing gradients, leading to unstable training. Thus we adopt the objective design by Wasserstein GAN:

\begin{equation}
\min_{G}\max_{||D||_{L} \leq 1}\mathbb{E}_{x\sim p_{\text{data}}(x)}[D(x)] - \mathbb{E}_{z\sim p_{\text{z}}(z)}[D(G(z))]
\end{equation}

The main differences between the Wasserstein GAN objective and the original one are 1) it is proved that optimizing the Wasserstein GAN objective is in fact minimizing the Wasserstein distance between two distributions; and 2) a Lipschitz constraint $||D||_{L} \leq 1$ is performed on the discriminator $D$ (also called a critic) to ensure the Wasserstein distance is solvable; and 3) the critic now learns to regress the Wasserstein distance between raw data distribution and synthetic data distribution. The Wasserstein distance, also known as the Earth Mover's Distance (EMD), is a distance function defined between two probability distributions:

\begin{equation}
\begin{aligned}
W_p(\mathcal{\Pr}_a, \mathcal{\Pr}_b)=(\min_{\gamma \in \Gamma} \sum_{i,j}\gamma(x_i,y_j)\|x_i-y_j\|_p)^\frac{1}{p} \\
s.t. \gamma 1 = \mathcal{\Pr}_a; \gamma^T 1 = \mathcal{\Pr}_b; \gamma\geq 0
\end{aligned}
\end{equation}

where $\Pr_{a}$ and $\Pr_{b}$ denote two probability distributions, $\Gamma$ denotes a set of all joint probability distributions whose marginals are $\Pr_{a}$ and $\Pr_{b}$. $\gamma \in \Gamma$ represents all transport plans to move $\Pr_{a}$ into $\Pr_{b}$ as well as the amount of mass they would take. Wasserstein distance measures the minimum cost of turning one probability distribution into another. The main advantage of using the Wasserstein distance is that it is a robust measure even when distributions do not align, which largely mitigates the gradient vanishing issues and prevents unstable training. Also, compared with the discriminator in the original GAN, the critic in Wasserstein GAN now regresses the distance between two distributions, which is a more straightforward design and suitable in our case, and the fitting quality is easier to assess. Note that, in CATS, we replace $x\sim p_{data}(x)$ with $x\sim p_{data}(x|\mathcal{M}^{stm})$ and $z\sim p_{z}(z)$ with $z\sim p_{z}(z|\mathcal{M}^{stm})$ to achieve data sampling conditional on $\mathcal{M}^{stm}$.

\section{Experiments and Results}

In this section, we introduce the evaluation experiments and performance results of the proposed CATS framework compared to other baseline approaches. First, we introduce the mobility dataset used and the experimental setup for the CATS framework. Then, we report the results from the Privacy Preservation (PP) tests and Characteristic Preservation (CP) tests to compare the effectiveness of privacy and spatiotemporal characteristic preservation between CATS and other baselines. Finally, we conduct Downstream Utility (DU) tests to demonstrate how CATS can be used to support common real-world applications, which further verifies the data utility of CATS.

\subsection{Dataset}

\begin{figure}[h]
  \centering
  \includegraphics[width=0.8\linewidth]{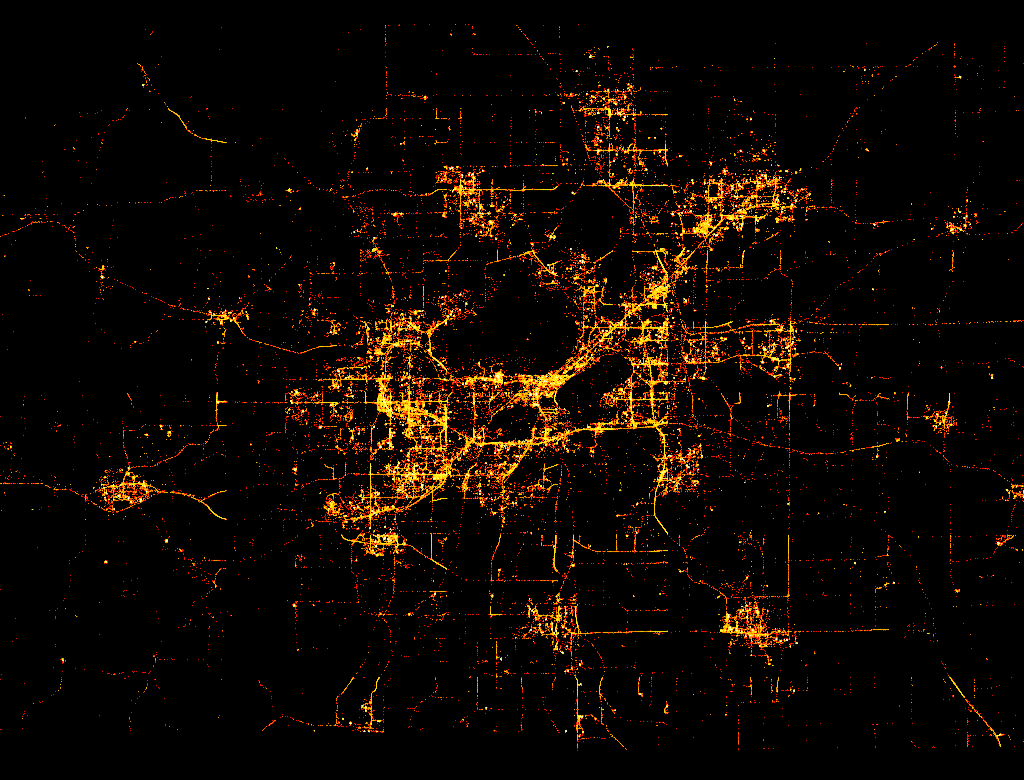}
  \caption{The overall geographic density distribution of the processed mobility data in Dane County, Wisconsin from September 7, 2020 to November 29, 2020.}
  \label{fig:overall_distribution}
\end{figure}

We use the UberMedia (acquired by Near in 2021, \url{https://near.com}) Mobility Dataset as our experiment data source. UberMedia Mobility Dataset records large-scale long-term individual GPS traces with high spatiotemporal resolution. The dataset provides both geographic coordinates (i.e., latitude and longitude) and timestamps (i.e., date, day of week, time of day) of trajectory points. Device IDs are also provided in the dataset as anonymized identifiers for users. 
We choose the Dane County, Wisconsin where the state capital city Madison locates as the research area and extract the data collected from September 7, 2020 to November 29, 2020 (i.e., 12 weeks). We then set the temporal resolution to 1 hour and extract the most active users from the dataset whose mobility is recorded continuously based on the following criteria: 1) users' movements are recorded over 80 days (i.e., over 95\% during the whole time period); and 2) users' movements are recorded at least 20 hours each day in a passive manner (i.e., users do not need to actively check-in their records). We then slice the whole records into daily trajectories. For a small portion (less than 1\%) of trajectories that has a few missing values, we interpolate them using the nearest active records. After preprocessing, 1,097 users with 90,395 dense daily trajectories containing 2,169,480 GPS records remain. Figure \ref{fig:overall_distribution} shows the overall geographic distribution of the dataset, where brighter pixels represent more active locations (i.e., containing more trajectory points). In our experiments, we further encode the trajectory points into a 128 × 128 squared grid system based on the size of study region, and the spatial resolution of the aggregated individual trajectory data is around $500$ m.

\subsection{Experimental Setup}
\paragraph*{Training Settings:} We implement the CATS via PyTorch, an open source deep learning framework. We split the users' trajectory dataset into a training set and a test set with a ratio of 4:1. Here we followed \cite{isola2018imagetoimage} and did not involve a static validation set as the dynamic and adversarial training process in CATS makes a static validation set not so helpful for hyperparameter tuning, which was also observed in our experiments. Since each user in the dataset has about 80 real daily trajectories on average and many of them are usually similar, we found that 64 synthetic trajectories are generally sufficient in simulating a user's mobility patterns from their real daily trajectories. Thus, we set the sample size of Conditional Sampling in CatGen to 64. Setting a lager sample size for synthetic trajectories are generally helpful but are more computationally expensive. The Location and Time Encodings are set to a dimension of 32 in our work, enabling the generation of learnable embeddings with a dimension greater than their original values (location is 2, and time is 24). This increase in dimensionality provides adequate space for embedding representation and has been explored in previous studies~\citep{may2020marc,rao2020lstm,mai2022review}. As the location and time embeddings are concatenated to form the spatiotemporal embedding, its dimension increases to 64, which also becomes the input dimension of the Global Context Attention and GRUs. The number of heads in the Global Context Attention is equivalent to the number of heads in its Multi-Head Self-Attention components, which is set to 8, following the default value by~\cite{vaswani2017attention}. For the training procedure, we optimize both CatGen and CatCrt for 50 epochs using the Root Mean Squared Propagation (RMSProp) with a learning rate of $2 \times 10^{-4}$. The remaining hyperparameters in the optimizer and training details follow the default settings established in~\cite{arjovsky2017wasserstein}. All the detailed hyperparameter setting in CATS can be found in the open source code repository on Github: \url{https://github.com/GeoDS/CATS}. 

\paragraph*{Baselines:} We compare our method with several common trajectory privacy protection approaches, namely Random Perturbation (RP), Gaussian Geomasking (GG), Differential Privacy (DP), and Trajectory K-Anonymization (TKA). For Differential Privacy, we refer to Geo-Indistinguishability, a location privacy standard extended from differential privacy, and we adopt the mechanism based on the controlled planar Laplacian noise that satisfies the $\epsilon$-Geo-Indistinguishability \citep{andres2013geo}. Depending on the scale at which planar Laplacian noise is applied, we further introduce two baselines: Location Differential Privacy (LDP) and Trajectory Differential Privacy (TDP). For Trajectory K-Anonymization (TKA), we introduce a variant where we group trajectories into clusters with a minimized size of $K$ and randomly assign trajectory points at each time step. Detailed settings of baselines are elaborated below.

\begin{itemize}
  \item Random Perturbation: It adds random noise following a uniform distribution $X \sim \mathcal{U}(a,b)$ to the latitude and longitude of each trajectory point in the dataset. $a$ and $b$ are lower bound and upper bound. In our experiment we set $a=-0.02$ and $b=0.02$.
  \item Gaussian Geomasking: It adds random noise following a gaussian distribution $X \sim \mathcal{N}(\mu,\,\sigma^{2})$ to the latitude and longitude of each trajectory point in the dataset. $\mu$ and $\sigma$ are mean and standard deviation. In our experiment we set $\mu=0$ and $\sigma=0.02$.
  \item Location Differential Privacy: It adds random noise following a planar Laplacian distribution $D_{\epsilon, x_0}(x)=\frac{\epsilon^2}{2\pi}e^{-\epsilon d(x_0, x)}$ to the latitude and longitude of each trajectory point in the dataset. $\epsilon$ is the scaling factor (also known as the privacy budget in standard differential privacy), $x_0$ is the center point of the distribution, $d(\cdot)$ is the distance between two points (e.g., Euclidean). In our experiment we set $\epsilon=100$ and $d(\cdot)$ using the Euclidean distance.
  \item Trajectory Differential Privacy: similar to LDP, but instead of adding planar Laplacian noise to each trajectory point, it treats each trajectory as a basic unit and add same noise to the points in each trajectory in batch.
  \item Trajectory K-Anonymization: we adopt the same location sampling strategy in CATS where we group trajectories to generate K-Anonymized spatiotemporal mobility matrics and perform conditional sampling. Then we randomly assign trajectory points at each time step to produce K-anonymized trajectories.
  \end{itemize}

\subsection{Characteristic Preservation Tests} \label{sec:PreservationTests}

\begin{figure}[H]
  \centering
  \includegraphics[width=\linewidth]{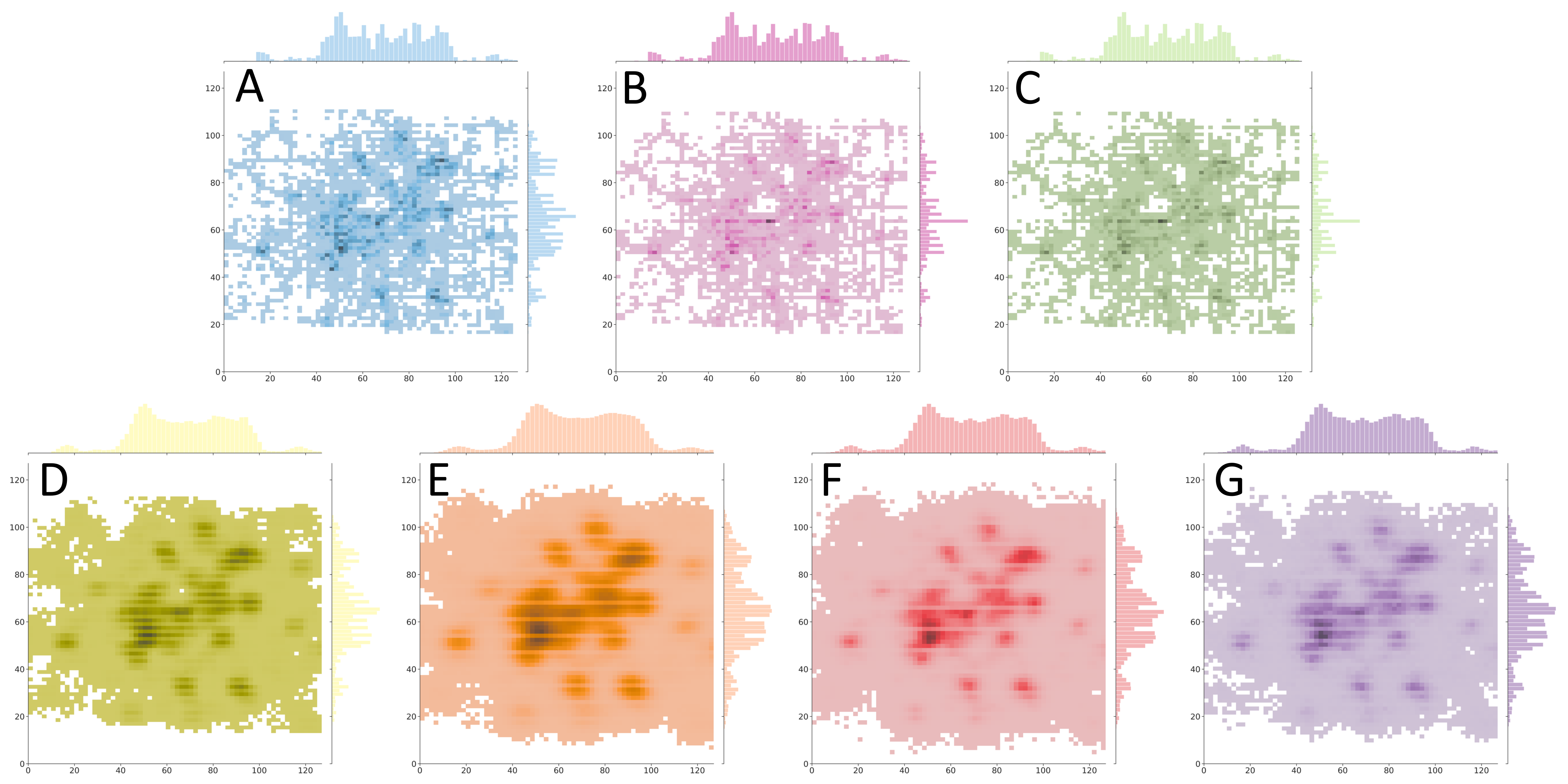}
  \caption{The geographic distributions of the raw trajectory dataset $\mathcal{X}_{raw}$ (A);  the trajectory dataset $\mathcal{X}_{tka}$ generated by TKA (B); the trajectory dataset $\mathcal{X}_{cats}$ generated by CATS (C); the trajectory dataset $\mathcal{X}_{rp}$ processed by RP (D); the trajectory dataset $\mathcal{X}_{gg}$ processed by GG (E); the trajectory dataset  $\mathcal{X}_{ldp}$ processed by LDP (F); and the trajectory dataset $\mathcal{X}_{tdp}$ processed by TDP (G). All the trajectory points are involved and encoded into a 128 x 128 grid system with a spatial resolution of 500 m. Note that the trajectories generated from CATS and TKA methods have the same aggregated geographic distribution since they are using the same location sampling strategy.}
  \label{fig:data_distribution}
\end{figure}

\begin{figure}[H]
  \centering
  \includegraphics[width=0.95\linewidth]{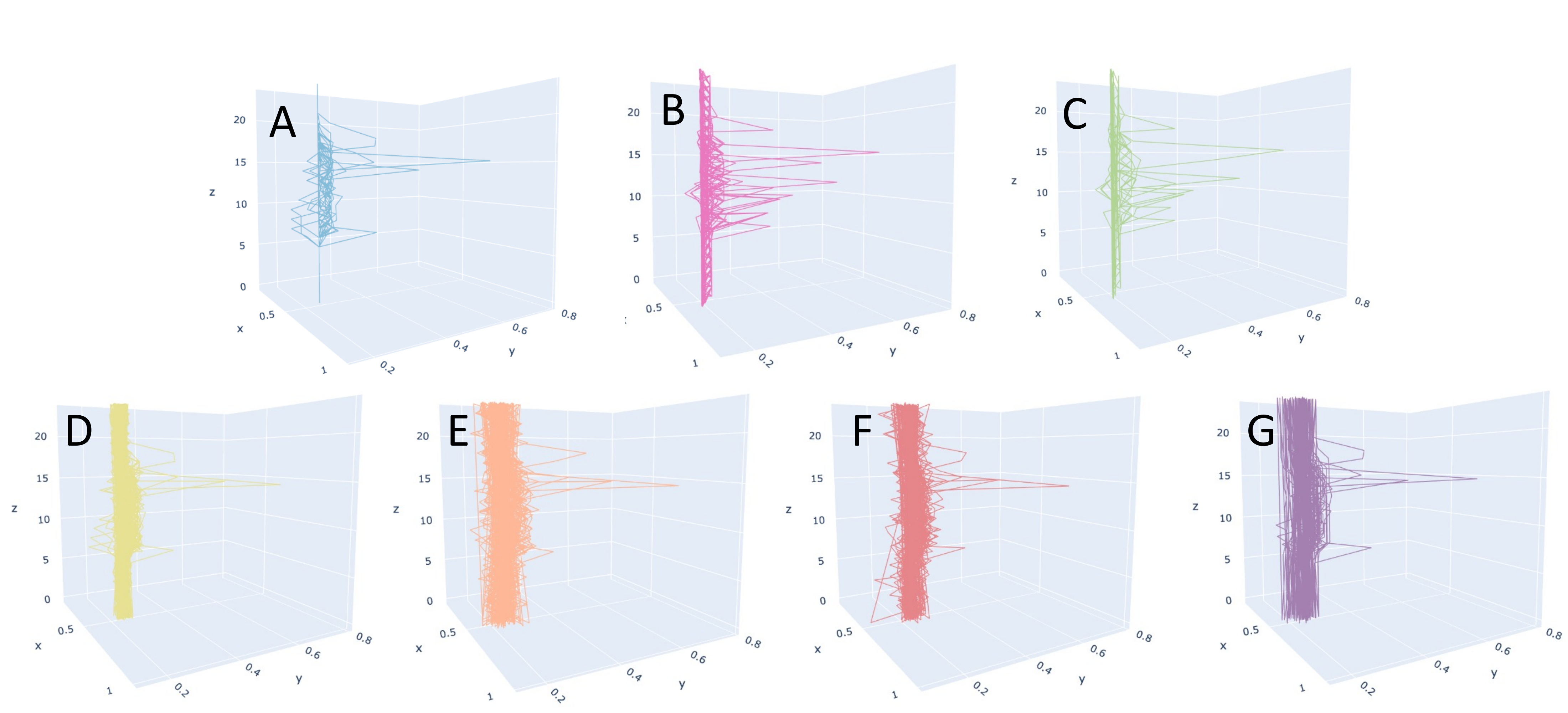}
  \caption{A visual comparison of individual-level trajectory data of the same user in a space-time cube. Sub-figure A comes from raw trajectory data $\mathcal{X}_{raw}$ and B-G come from generated or processed trajectory datasets $\mathcal{X}_{tka}$, $\mathcal{X}_{cats}$, $\mathcal{X}_{rp}$, $\mathcal{X}_{gg}$, $\mathcal{X}_{ldp}$, $\mathcal{X}_{tdp}$, respectively.}
  \label{fig:individual_examples}
\end{figure}

\begin{figure}[h]
  \centering
  \includegraphics[width=0.5\linewidth]{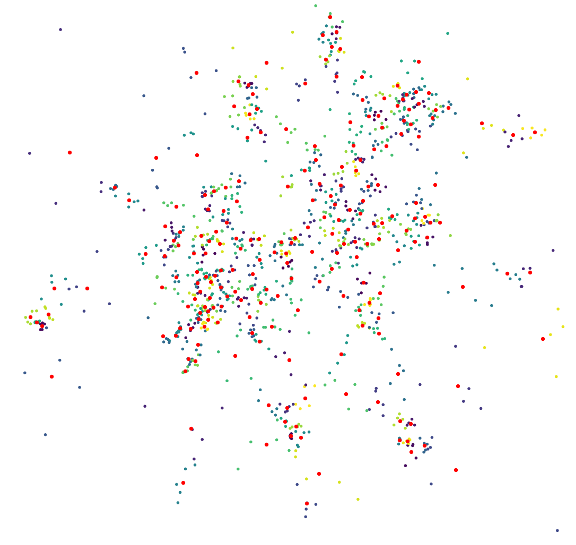}
  \caption{Constrained K-Means user clustering result on $\mathcal{X}_{raw}$ using users' trajectory centroids. Larger red dots represent for cluster centers, and the neighboring smaller dots with the same color represent a user cluster (i.e., a group of user’s centroids)}.
  \label{fig:constrained_K_means}
\end{figure}

In this part, we analyze and compare the spatiotemporal data characteristics of the raw trajectory dataset (denoted as $\mathcal{X}_{raw}$) and the trajectory datasets processed by RP, GG, LDP, TDP, TKA and our method (denoted as $\mathcal{X}_{rp}$, $\mathcal{X}_{gg}$, $\mathcal{X}_{ldp}$, $\mathcal{X}_{tdp}$, $\mathcal{X}_{tka}$ and $\mathcal{X}_{cats}$, respectively). The geographic distribution of these datasets can be seen in Figure \ref{fig:data_distribution}. A visual comparison of individual-level trajectory data of the same user processed from different methods is also provided in Figure \ref{fig:individual_examples}. Specifically, $\mathcal{X}_{cats}$ is generated from the K-anonymized spatiotemporal mobility matrices ($k=5$) derived from the KAMA module in CATS. It can be seen that, for overall geographic distributions, the result from our method (Figure \ref{fig:data_distribution}C) and TKA (with the same location sampling strategy as CATS, Figure \ref{fig:data_distribution}B) are more visually similar with the raw geographic distribution (Figure \ref{fig:data_distribution}A). For the individual-level trajectory, the result of our method (Figure \ref{fig:individual_examples}C) preserves patterns more similar to the raw trajectory (Figure \ref{fig:individual_examples}A) than that of other methods. It is also noteworthy that our method and TKA generate synthetic trajectories based on k-anonymized mobility of users where each location was visited in real trajectories, providing an additional advantage over noise-based baselines which may create less realistic and meaningful trajectory points in a constrained urban space. Figure \ref{fig:constrained_K_means} shows the constrained clustering result on $\mathcal{X}_{raw}$. Larger red dots represent for cluster centers, and the neighboring smaller dots with the same color represent a user cluster (i.e., a group of user's centroids). We design two Characteristic Preservation (CP) tests. The first test (i.e., CP1) examines both collective-level and individual-level mobility measures of trajectory datasets to investigate to what extent the processed datasets can preserve collective spatiotemporal distribution and individual properties. The second test (i.e., CP2) performs the origin-destination demand estimation derived from the trajectories to investigate to what extent the processed datasets can preserve movement flows between locations. The selection of mobility measures in CP tests is motivated by their widespread use in mobility research and their potential to enable a comprehensive assessment of spatiotemporal characteristics in trajectories. Furthermore, the majority of these measures are readily available in Scikit-Mobility~\citep{JSSv103i04}, a standardized computation library for human mobility analysis in Python. This feature facilitates a reliable benchmark for evaluating the performance of our approach against existing methods.

\subsubsection{CP1: Mobility Measures}

We first investigate collective-level mobility measures via the Wasserstein distance, Jensen–Shannon divergence, and entropy. We use the Wasserstein distance of order two ($W_2$) to measure the distances between spatiotemporal distributions of trajectory datasets and introduce two specific measures: the Overall Wasserstein Distance, which measures the $W_2$ distance between overall distributions of trajectory points in two datasets; and the Time-Specific Wasserstein Distance, which measures a series of $W_2$ distances between hourly distributions of trajectory points in two datasets. These two measures can tell the differences between two datasets both at the overall scale and at different time scales. Please note that the Wasserstein distance calculated here as a mobility measure is different from that of optimization (in section \ref{sec:optimization}) which mainly estimates the distance between the probability distributions of how raw or synthetic trajectory points are connected (i.e., trajectory topology) in a high dimensional space.

In addition, we use Jensen–Shannon divergence (JSD), a symmetric metric of measuring the similarity between two probability distributions, to measure the overall and time-specific distances between spatiotemporal distributions of trajectory datasets. The definition of JSD is defined as below:

\begin{equation}
{\rm {D_{\text{JS}}}}(\mathcal{\Pr}_{a}\parallel \mathcal{\Pr}_{b})={\frac {1}{2}}D_{\text{KL}}(\mathcal{\Pr}_{a}\parallel \mathcal{\Pr}_{m})+{\frac {1}{2}}D_{\text{KL}}(\mathcal{\Pr}_{b}\parallel \mathcal{\Pr}_{m}),
\end{equation}

where $\mathcal{\Pr}_{a}$ and $\mathcal{\Pr}_{b}$ denote two two-dimensional probability distributions; $D_{\text{JS}}$ demotes the JSD between $\mathcal{\Pr}_{a}$ and $\mathcal{\Pr}_{b}$; $\mathcal{\Pr}_{m}={\frac  {1}{2}}(\mathcal{\Pr}_{a}+\mathcal{\Pr}_{b})$ is the average probability distribution of $\mathcal{\Pr}_{a}$ and $\mathcal{\Pr}_{b}$; $D_{\text{KL}}$ is the Kullback–Leibler divergence (KLD), a non-symmetric distance between two probability distributions. For example, $D_{\text{KL}}(\mathcal{\Pr}_{a}\parallel \mathcal{\Pr}_{m})$ is defined as below:

\begin{equation}
D_{\text{KL}}(\mathcal{\Pr}_{a}\parallel \mathcal{\Pr}_{m})=\sum _{x, y}\mathcal{\Pr}_{a}(x, y)\log \left({\frac {\mathcal{\Pr}_{a}(x, y)}{\mathcal{\Pr}_{m}(x, y)}}\right)
\end{equation}

Another entropy-based collective measure we use is the Random Location Entropy (denoted as $LE_{rand}$), which is defined as below:

\begin{equation}
LE_{rand}(l_i) = log_2(N_{l_i})
\end{equation}

where $N_{l_i}$ is the number of distinct individuals that visited location $l_i$. $LE_{rand}(l_i)$ captures the degree of predictability of $l_i$ if each individual visits it with equal probability.

We also investigated individual-level mobility measures with regard to entropy and geometric properties. We first introduce three entropy-based measures, which are important indicators for understanding the predictability in human mobility \citep{song2010limits}: Random Entropy (denoted as $E_{rand}$, capturing the degree of predictability of a user $u$ whereabouts if each location is visited with equal probability), Temporal-Uncorrelated Entropy (denoted as $E_{unc}$, characterizing the heterogeneity of visitation patterns of a user $u$), and Actual Entropy (denoted as $E_{act}$, capturing the full spatiotemporal order presented in the mobility pattern of a user $u$). Their formal definitions are as follows:

\begin{equation}
E_{rand}(u) = log_2(N_u)
\end{equation}
\begin{equation}
E_{unc}(u) = - \sum_{l_i=1}^{N_u} p_u(l_i) log_2 p_u(l_i)
\end{equation}
\begin{equation}
E_{act}(u) = - \sum_{T'^u \in T^u}P(T'^u)log_2[P(T'^u)]
\end{equation}

where $N_u$ denotes the number of distinct locations visited by user $u$, $p_u(l_i)$ denotes the historical probability that location $l_i$ was visited by user $u$, $T^u$ denotes a trajectory by user $u$, and $P(T'^u)$ denotes the probability of finding a particular time-ordered subsequence $T'^u$ in the trajectory $T^u$. Note that $E_{rand} \geqslant E_{unc} \geqslant E_{act}$ for each user.

We then introduce four geometry-based measures: Jump Length (denoted as $\Delta l$, calculating the total geographic distance between every two consecutive locations in a trajectory by user $u$), Location Switch (denoted as $N_{ls}$, counting the total times of location changes between every two consecutive locations in a trajectory by user $u$), Radius of Gyration (denoted as $r_g$, computing the radius of gyration that indicates the characteristic distance travelled by user $u$), and Azimuth-Based Tortuosity (denoted as $Tor$, describing the average direction change in a trajectory by user $u$). Their formal definitions are as follows:


\begin{equation}
\Delta l(T^u) = \sum_{l_i \in T^u} Haversine(l_i, l_{i + 1})
\end{equation}
\begin{equation}
N_{ls}(T^u) = \sum_{l_i \in T^u} \mathbbm{1}(l_i \neq l_{i+1})
\end{equation}
\begin{equation}
r_g(u) = \sqrt{ \frac{1}{N_l^u} \sum_{i=1}^{N_l^u} Haversine(l_i, l_c^u)^2}
\end{equation}
\begin{equation}
Tor(T^u) = \frac{1}{|\{l_i \in T^u\}|}\sum_{l_i \in T^u} |Azimuth(l_i, l_{i + 1}) - Azimuth(l_{i-1}, l_{i})|
\end{equation}
where $N_l^u$ is the total number of locations in the trajectory data by user $u$, $l_c^u$ is the location centroid of the trajectory data of user $u$, $Haversine$ and $Azimuth$ are the functions for calculating the Haversine distance and Azimuth angle between two locations, respectively. 

\begin{table}[h]
	\centering
	\caption{The collective-level and individual-level mobility measures of raw trajectory dataset and privacy-preserved trajectory datasets.}
\renewcommand{\arraystretch}{1.6}
\begin{tabular}{c|c|cccccc}\hline
 \multicolumn{8}{c}{\textbf{Collective-level Mobility Measures}} \\\hline
 \multirow{2}{*}{\textbf{Measures (mean)}} & \multicolumn{6}{c}{\textbf{Datasets}}\\\cline{2-8}
 & \textbf{$\mathcal{X}_{raw}$} & \textbf{$\mathcal{X}_{rp}$} & \textbf{$\mathcal{X}_{gg}$} & \textbf{$\mathcal{X}_{ldp}$} & \textbf{$\mathcal{X}_{tdp}$} & \textbf{$\mathcal{X}_{tka}$}  & \textbf{$\mathcal{X}_{cats}$} (ours)\\\hline
 Overall $W_{2}$ Distance   &0.000	&6.096	&11.554	&9.529	&9.607 &1.181 &\textbf{1.181}\\\hline
 Overall JSD   & 0.000 & 0.283	& 0.320 & 0.279 & 0.280 &0.247 & \textbf{0.247} \\\hline
 Time-Specific $W_{2}$ Distance   &0.000	&2.528	&5.909	&4.114	&4.161 &1.483 &\textbf{1.483}\\\hline
 Time-Specific JSD  & 0.000 & 0.311	& 0.349	& 0.306	& 0.306 &0.267 & \textbf{0.267}\\\hline
 Random Location Entropy    &1.003	&2.860	&3.797	&3.712  &2.854 &2.830	&\textbf{2.830}\\\hline
 \multicolumn{8}{c}{\textbf{Individual-level Mobility Measures}} \\\hline
 \multirow{2}{*}{\textbf{Measures (mean)}} & \multicolumn{6}{c}{\textbf{Datasets}}\\\cline{2-8}
 & \textbf{$\mathcal{X}_{raw}$} & \textbf{$\mathcal{X}_{rp}$} & \textbf{$\mathcal{X}_{gg}$} & \textbf{$\mathcal{X}_{ldp}$} & \textbf{$\mathcal{X}_{tdp}$} & \textbf{$\mathcal{X}_{tka}$} & \textbf{$\mathcal{X}_{cats}$} (ours)\\\hline
 Random Entropy    &5.712	&10.949	&10.949	&10.949  &7.973 &6.235	&\textbf{6.235}\\\hline
Temporal-Uncorrelated Entropy    &1.359	&10.949	&10.949	&10.949	&7.174 &3.161	&\textbf{3.161}\\\hline
Actual Entropy    &0.648	&10.944	&10.944	&10.944	&2.529 &1.516	&\textbf{1.516}\\\hline
Jump Length (km)   &0.428	&1.394	&2.105	&1.813	&\textbf{0.428} &1.961	&1.046\\\hline
Location Switch    &3.823	&23.000	&23.000	&23.000	&\textbf{3.823} &16.286	&10.866\\\hline
Radius of Gyration (km) &2.051	&\textbf{2.298} &2.655	&2.532	&2.523 &2.361 &2.361\\\hline
Azimuth-Based Tortuosity    &5.218	&31.390	&31.390	&31.390	&\textbf{5.218} &22.228	&14.832\\\hline
\end{tabular}
	\label{table:cp1_mobility_measure}
\end{table}

\begin{figure}[!]
  \centering
  \includegraphics[width=\linewidth]{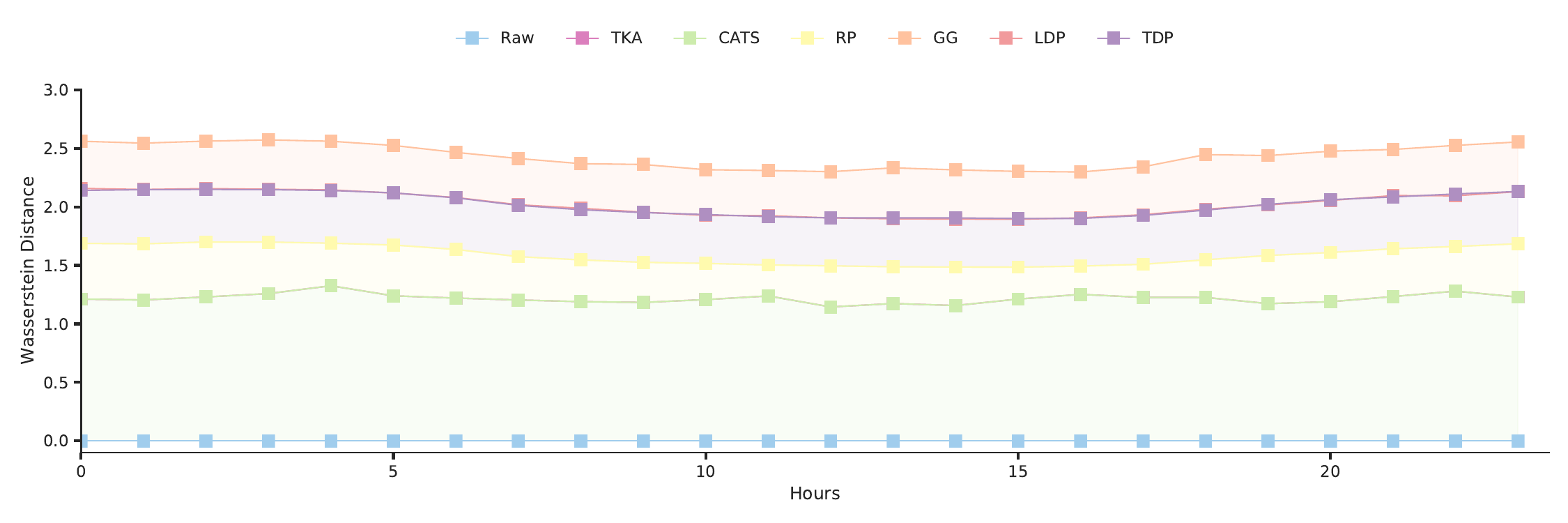}
  \caption{The Time-Specific Wasserstein Distance to $\mathcal{X}_{raw}$ at each hour. Note that the trajectories generated from CATS and TKA methods have the same aggregated geographic distribution and time-specific $W_2$ distance to raw trajectory data.}
  \label{fig:time_specific_w_dist}
\end{figure}

The results of both collective-level and individual-level mobility measures can be found in Table \ref{table:cp1_mobility_measure}. Here we report the mean value for each measure to make them easy to compare. For the Time-Specific Wasserstein distance, we also show how it changes over time (i.e., hourly) in Figure \ref{fig:time_specific_w_dist}. From the collective perspective, compared with other datasets, $\mathcal{X}_{cats}$ (and also $\mathcal{X}_{tka}$, which adopts the same location sampling strategy of $\mathcal{X}_{cats}$ and thus has the same aggregated geographic distribution) shows both the lowest overall and time-specific $W_{2}$ distance and JSD to $\mathcal{X}_{raw}$ and the most similar $LE_{rand}$ to $\mathcal{X}_{raw}$, indicating better similarity to $\mathcal{X}_{raw}$ in spatiotemporal distribution and location entropy. From the individual perspective, compared with other methods, $\mathcal{X}_{cats}$ shares more similarity with $\mathcal{X}_{raw}$ in all the three entropy-based properties and the second best in one geometric property: radius of gyration. Trajectory differential privacy approach $\mathcal{X}_{tdp}$ shares better similarity with $\mathcal{X}_{raw}$ in most of geometric properties including location switch, jump length, and azimuth-based tortuosity.

\subsubsection{CP2: Origin-Destination Demand Estimation}

\begin{figure}[h]
  \centering
  \includegraphics[width=\linewidth]{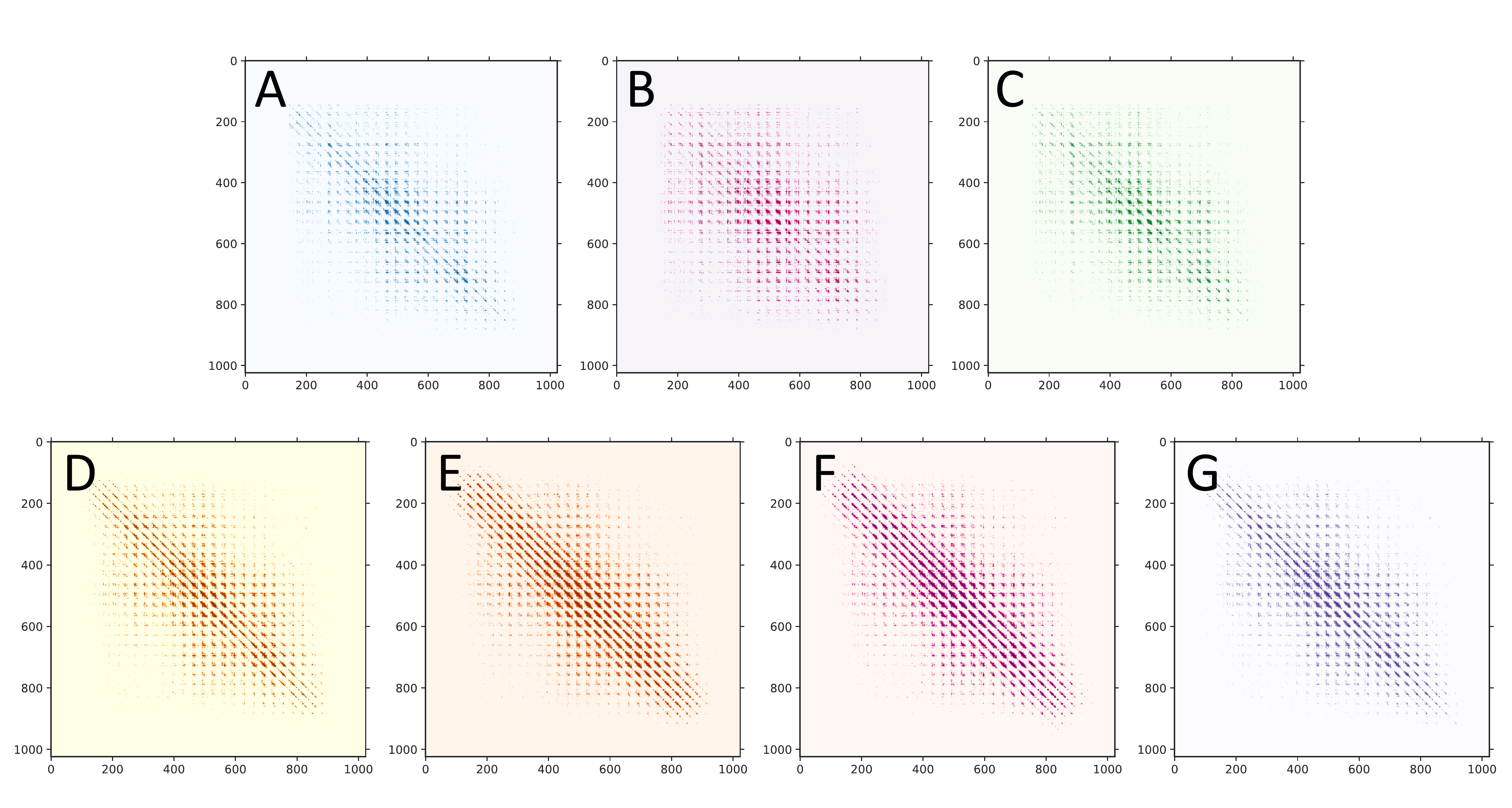}
  \caption{The comparison of origin-destination flow matrices (with logarithmic values plotted) derived from different trajectory datasets: A: $M^{od}_{\mathcal{X}_{raw}}$, B: $M^{od}_{\mathcal{X}_{tka}}$, C: $M^{od}_{\mathcal{X}_{cats}} (ours)$, D: $M^{od}_{\mathcal{X}_{rp}}$, E: $M^{od}_{\mathcal{X}_{gg}}$, F: $M^{od}_{\mathcal{X}_{ldp}}$, and G: $M^{od}_{\mathcal{X}_{tdp}}$.}
  \label{fig:od_flow_matrix}
\end{figure}

In light of the spatial extent and land use of the research area (i.e., Madison, WI), a spatial resolution of 32x32 regions (1,024 regions in total) was deemed to be sufficient for origin-destination demand estimation in the study region~\citep{hou2021intracounty}. We evenly divide the whole research area into $32 \times 32$ regions and calculate the origin-destination mobility flow matrix (OD matrix) based on hourly trips among these regions in each privacy-preserved trajectory dataset (denoted as $M^{od}_{\mathcal{X}_{rp}}$, $M^{od}_{\mathcal{X}_{gg}}$, $M^{od}_{\mathcal{X}_{ldp}}$, $M^{od}_{\mathcal{X}_{tdp}}$, $M^{od}_{\mathcal{X}_{tka}}$, and $M^{od}_{\mathcal{X}_{cats}}$, respectively) and investigate their Structural Similarity (SSIM)~\citep{wang2004image} to the OD matrix of raw dataset (denoted as $M^{od}_{\mathcal{X}_{raw}}$). SSIM is a quantitative measure used to compare the similarity of two images. Recent studies also use SSIM to compare two OD flow matrices because OD pairs are analogous to the pixels of an image~\citep{behara2021geographical}. The visualization of each OD matrix (the logarithmic values plotted for better visual presentation) can be found in Figure \ref{fig:od_flow_matrix}. The SIMM values with different Gaussian kernels and the mean SIMM values are reported in Table \ref{table:ssim}. As shown in Table \ref{table:ssim}, the OD matrix of $\mathcal{X}_{cats}$ has the highest SSIM scores both at three different scales and at the average level, indicating better overall similarity to $\mathcal{X}_{raw}$ in preserving movement patterns and transport demands between locations. Such similarity can also be visually observed in Figure \ref{fig:od_flow_matrix}.

\begin{table}[!htb]
	\centering
	\caption{The structural similarity (SSIM) between the OD matrix of original trajectory dataset ($M^{od}_{\mathcal{X}_{raw}}$) and the OD matrices of privacy-preserved trajectory datasets ($M^{od}_{\mathcal{X}_{rp}}$, $M^{od}_{\mathcal{X}_{gg}}$, $M^{od}_{\mathcal{X}_{ldp}}$, $M^{od}_{\mathcal{X}_{tdp}}$, $M^{od}_{\mathcal{X}_{tka}}$, and $M^{od}_{\mathcal{X}_{cats}}$). Sigma is the standard deviation for the normalized Gaussian kernel in SSIM, determining different window sizes.}
\renewcommand{\arraystretch}{1.6}
\begin{tabular}{@{}ccccccc@{}}
 \hline
 \textbf{SIMM} & \textbf{$M^{od}_{\mathcal{X}_{rp}}$} & \textbf{$M^{od}_{\mathcal{X}_{gg}}$} & \textbf{$M^{od}_{\mathcal{X}_{ldp}}$} & \textbf{$M^{od}_{\mathcal{X}_{tdp}}$} & \textbf{$M^{od}_{\mathcal{X}_{tka}}$}  & \textbf{$M^{od}_{\mathcal{X}_{cats}}$} (ours)\\\hline
 Sigma=128   &0.358	&0.157	&0.281	&0.889	&0.766	&\textbf{0.916}\\\hline
 Sigma=96    &0.356	&0.155	&0.279	&0.885	&0.762	&\textbf{0.914}\\\hline
 Sigma=64    &0.420	&0.247	&0.356	&0.876	&0.759	&\textbf{0.917}\\\hline
 Average    &0.378	&0.186	&0.305	&0.883	&0.762	&\textbf{0.916}\\\hline
\end{tabular}
	\label{table:ssim}
\end{table}

\subsection{Privacy Preservation Tests}

In this part, we analyze and compare the effectiveness of privacy protection of RP, GG, LDP, TDP, TKA, and our method CATS. We design two Privacy Preservation (PP) tests (also known as attacks) covering two main trajectory privacy topics: user identity and home location detection. The first test (i.e., PP1) examines the Trajectory-User Linking (TUL)~\citep{gao2017identifying} to investigate to what extent the privacy-preserved datasets can prevent users' identity from being revealed. The second test (i.e., CP2) performs the Home Location Clustering (HLC)~\citep{gao2019exploring} to investigate to what extent the privacy-preserved datasets can prevent users' home locations from being revealed.

\subsubsection{PP1: Trajectory-User Linking}

Trajectory-User Linking (TUL) is a classic LBS task that tries to link anonymous trajectories to their creators (i.e., users)~\citep{gao2017identifying}. Many attempts are made in recent studies to utilize deep learning or machine learning models to re-identify the users of public trajectories and reveal their location preferences and social circles, which raises public concerns of privacy. We utilize the Multiple-Aspect tRajectory Classifier (MARC)~\citep{may2020marc}, and design a deep-learning-based trajectory classification model to perform the trajectory-user linking task on the privacy-preserved trajectory dataset to test if they can protect users' identity. We randomly split $\mathcal{X}_{raw}$ into a training set, a validation set, and a test set based on trajectory IDs (with the ratio of 3:1:1), and then we apply the same splitting strategy on $\mathcal{X}_{rp}$, $\mathcal{X}_{gg}$, $\mathcal{X}_{ldp}$, $\mathcal{X}_{tdp}$, $\mathcal{X}_{tka}$, and $\mathcal{X}_{cats}$. We train the model on the training set of $\mathcal{X}_{raw}$ for 30 epochs (with a learning rate 0.001 and an embedding size 100) and choose the model weight that has the best performance on the validation set. Then, we run the trained model on all the test sets and evaluate the model performance with five TUL measures: Macro Precision (the mean precision among all users), Macro Recall (the mean recall among all users), Macro F1 score (the harmonic mean of Macro Precision and Macro Recall), Top-1 Accuracy (the accuracy to have the correct user to be the most probable user candidate), and Top-5 Accuracy (the accuracy to have the correct user among the top-5 most probable user candidates). The lower the TUL measures, the lower possibility the users' identity can be revealed from trajectory data, and thus the higher effectiveness of user identity protection. The TUL measures on test sets of each trajectory dataset are reported in Table \ref{table:tulmetrics}. Because of the K-anonymity guaranteed by the constrained K-Means clustering on spatiotemporal mobility matrices in our methodological framework, both $\mathcal{X}_{cats}$ and $\mathcal{X}_{tka}$ achieve the comparable and lowest scores in all the TUL measures, indicating better effectiveness of user identity protection.

\begin{table}[h]
	\centering
	\caption{The Trajectory-User Linking measures on the test set of raw trajectory dataset and privacy-preserved trajectory datasets.}
\renewcommand{\arraystretch}{1.6}
{\begin{tabular}{c|c|cccccc}
 \hline
\textbf{TUL Measures} & \textbf{$\mathcal{X}_{raw}$} & \textbf{$\mathcal{X}_{rp}$} & \textbf{$\mathcal{X}_{gg}$} & \textbf{$\mathcal{X}_{ldp}$} & \textbf{$\mathcal{X}_{tdp}$} & \textbf{$\mathcal{X}_{tka}$}  & \textbf{$\mathcal{X}_{cats}$} (ours)\\\hline
 Macro Precision    &0.801		&0.730		&0.595		&0.617		&0.544		&0.430		&\textbf{0.425}\\\hline
 Macro Recall       &0.783		&0.721		&0.576		&0.601		&0.532		&\textbf{0.408}		&0.411\\\hline
 Macro F1 Score     &0.775		&0.698		&0.569		&0.594		&0.518		&\textbf{0.388}		&0.393\\\hline
 Top-1 Accuracy     &0.812		&0.743		&0.620		&0.673		&0.532		&0.427		&\textbf{0.421}\\\hline
 Top-5 Accuracy     &0.951		&0.850		&0.807		&0.833		&0.767		&\textbf{0.649}		&0.656\\\hline
\end{tabular}}
	\label{table:tulmetrics}
\end{table}

\subsubsection{PP2: Home Location Clustering}

We use the Density-Based Spatial Clustering of Applications with Noise (DBSCAN), a common location clustering algorithm to detect users' home locations from trajectory data~\citep{gao2019exploring}. We identify nighttime as 8 PM - 7 AM next day and define the location clusters detected during nighttime are possible home location clusters. Generally, the larger the cluster, the more possible it covers the home location of the user. We set the minimum number of points (MinPts) to 4 and set the search radius (i.e., eps) in a range of $0.002-0.042$ degrees (roughly 200 m - 4,600 m) in steps of 0.002 degree. We run the home location clustering on all the trajectory datasets with the same settings and evaluate three HLC measures: Centroid Shift, Medoid Shift, and Number of Home Clusters. The Centroid Shift and Medoid Shift both measure the shift distance between representative centers of home location clusters detected in raw trajectory dataset and privacy-preserved trajectory datasets. The Number of Home Clusters indicates the number of home location candidates, which also implies a level of k-anonymity with regard to home locations in each dataset. The higher the HLC measures, the lower possibility the users' home locations can be revealed from trajectory data, and thus the higher effectiveness of user home location protection. The HLC measures on each trajectory dataset are reported in Table \ref{table:hlc_metrics}. Here we report the mean and median values for each measure at $eps=0.02$ degree to make them easy to compare. 
In short, both $\mathcal{X}_{cats}$ and $\mathcal{X}_{tka}$ achieve the highest scores in most of the HLC measures, indicating better effectiveness of user home location protection. But $\mathcal{X}_{cats}$ performs better than $\mathcal{X}_{tka}$ in the individual-level mobility measures and OD demand estimation (See section \ref{sec:PreservationTests}).

\begin{table}[h]
	\centering
	\caption{The Home Location Clustering measures (in degree) on raw trajectory dataset and privacy-preserved trajectory datasets.}
\renewcommand{\arraystretch}{1.6}
\begin{tabular}{c|c|cccccc}
 \hline
 \textbf{HLC Measures} & \textbf{$\mathcal{X}_{raw}$} & \textbf{$\mathcal{X}_{rp}$} & \textbf{$\mathcal{X}_{gg}$} & \textbf{$\mathcal{X}_{ldp}$} & \textbf{$\mathcal{X}_{tdp}$} & \textbf{$\mathcal{X}_{tka}$} & \textbf{$\mathcal{X}_{cats}$} (ours)\\\hline
 Mean Centroid Shift   &0.000	&0.007	&0.010	&0.010	&0.022 &0.024 &\textbf{0.024}\\\hline
 Median Centroid Shift    &0.000	&0.001	&0.001  &0.001	&0.003 &0.010	&\textbf{0.010}\\\hline
 Mean Medoid Shift    &0.000	&0.008	&0.010	&0.009  &0.022 &0.025	&\textbf{0.025}\\\hline
 Median Medoid Shift    &0.000	&0.001	&0.001	&0.001  &0.003 &0.010	&\textbf{0.010}\\\hline
 Mean Number of Home Clusters    &1.946	&1.564	&1.332	&1.389  &\textbf{3.520} &3.106	&3.106\\\hline
 Median Number of Home Clusters    &2.000	&1.000	&1.000	&1.000  &\textbf{3.000} &\textbf{3.000}	&\textbf{3.000}\\\hline
\end{tabular}
	\label{table:hlc_metrics}
\end{table}

\subsection{Downstream Utility Tests}

The experiments and results of CP tests and PP tests reported above indicate that our method can better preserve spatiotemporal characteristics and privacy of trajectory datasets compared with other baselines, thus promising for privacy-preserving trajectory data publication. In this part, we further verify the data utility of the method in selected downstream tasks in GIScience. We design two Downstream Utility (DU) tests containing two common real-world downstream applications with trajectory data: Location Recommendation and Trajectory Reconstruction. The first test (i.e., DU1) performs a location recommendation task based on privacy-preserved trajectory datasets to examine if our method can achieve competitive accuracy. Specifically, DU1 represents an application scenario where the raw trajectory dataset exists and we want to publish a privacy-preserved version with high data utility. The second test (i.e., DU2) performs trajectory reconstruction based on the overall spatiotemporal distribution of mobility statistics in a geographic region. Differing from DU1, DU2 represents another application scenario where the raw trajectory dataset does not exist and we want to reconstruct a trajectory dataset following the macroscopic spatiotemporal statistics with basic data utility.

\subsubsection{DU1: Location Recommendation}

In DU1, we formulate the recommendation task as a common task in recommendation system: the Click-Through Rate (CTR) prediction problem, in which we attempt to predict the probability that a user will click on a recommended item. In the context of location recommendation, we predict the probability that a user will visit a recommended next location. Specifically, we use a Factorization Machine (FM) to solve the CTR prediction problem for location recommendation \citep{rao2021privacy}. The definition of FM is as below:

\begin{equation}
\hat{y} = w_{0} + \sum_{i=1}^{n} w_{i}x_{i} + \sum_{i=1}^{n} \sum_{j=i+1}^{n} <v_{i}, v_{j}> x_{i}x_{j}
\end{equation}

Where $x_{i}$ is the i-th feature variable, $w_{0}$ is the bias item, $w_{i}$ is the weight of the i-th feature variable (order-1 feature interactions), and $<v_{i}, v_{j}>$ is the weight of the interaction between the i-th and j-th feature variables (order-2 feature interactions), where $v_{i}$ and $v_{j}$ are the latent feature vectors of the i-th and j-th feature variables, respectively. FM is able to capture higher-order feature interactions by considering the inner product of the feature latent vectors.

\begin{figure}[h]
  \centering
  \includegraphics[width=\linewidth]{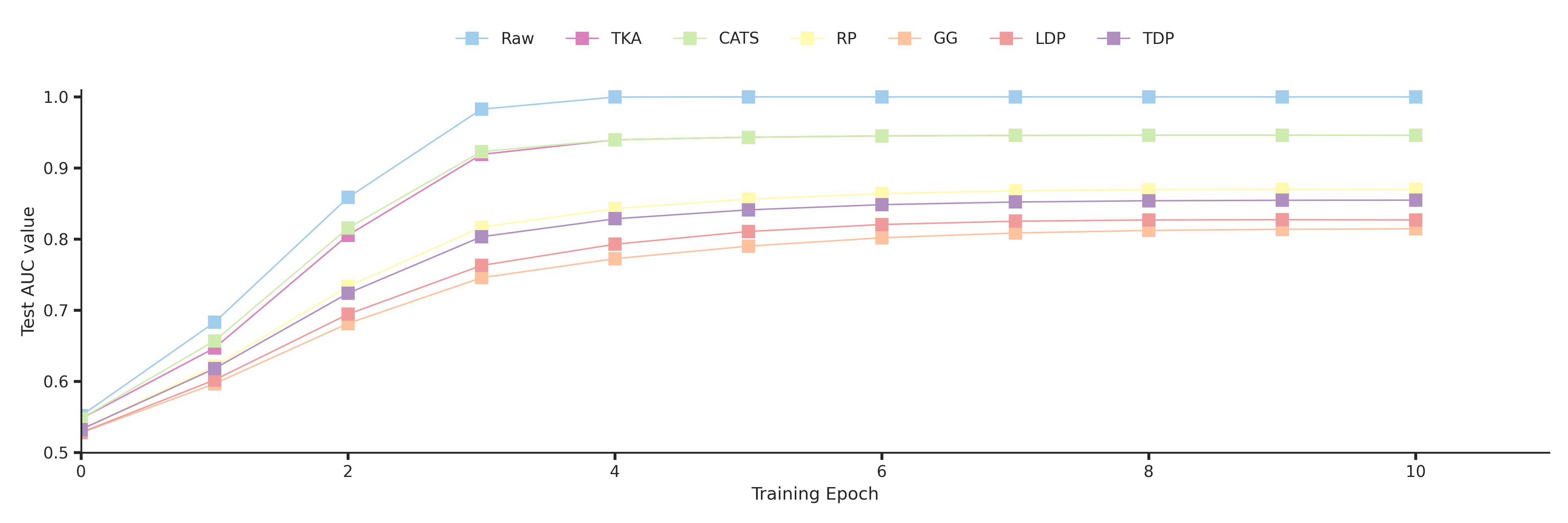}
  \caption{The comparison of location recommendation accuracy by AUC curves on the test sets of $\mathcal{X}_{raw}$, $\mathcal{X}_{rp}$, $\mathcal{X}_{gg}$, $\mathcal{X}_{ldp}$, $\mathcal{X}_{tdp}$, $\mathcal{X}_{tka}$, and $\mathcal{X}_{cats}$.}
  \label{fig:loc_rec_test_auc}
\end{figure}

We train FM on the training sets of $\mathcal{X}_{raw}$, $\mathcal{X}_{rp}$, $\mathcal{X}_{gg}$, $\mathcal{X}_{ldp}$, $\mathcal{X}_{tdp}$, $\mathcal{X}_{tka}$, and $\mathcal{X}_{cats}$ respectively, and run the trained FM model on the test set of $\mathcal{X}_{raw}$ to evaluate location recommendation accuracy. The accuracy metric we use is area under the receiver operating characteristic (AUC). In this case, the higher the AUC value, the better the location recommendation accuracy. The comparison of test AUC curves between the raw trajectory dataset and privacy-preserved trajectory datasets is shown in Figure \ref{fig:loc_rec_test_auc}. We can see that $\mathcal{X}_{cats}$ has the highest test AUC value (0.945) over all the other privacy-preserved trajectory datasets while the AUC value of $\mathcal{X}_{raw}$ is close to 1, indicating high data utility in location recommendation.

\subsubsection{DU2: Trajectory Reconstruction}

In DU2, we demonstrate how to reconstruct daily trajectory data with our method from sparse social-media geotagged check-in data. Due to the nature of social media, the recording of users' check-in locations is usually discontinuous with irregular time intervals, resulting in data sparsity and irregularity. Nevertheless, the statistical-level mobility patterns such as location visit frequency and origin-destination transportation demands are still preserved in the spatiotemporal distribution of check-in data \citep{hawelka2014geo}. By leveraging CATS, here we conduct a case study in New York City (NYC) to reconstruct users' trajectories from statistical-level mobility patterns obtained from social-media geotagged check-in data.

We use the data from the NYC Foursquare social media check-in dataset collected from April 2012 to February 2013, which is extracted and shared by \cite{may2020marc,yang2014modeling}. After data cleaning and preprocessing, there are 193 users with 66,962 time-ordered check-in records in the dataset, containing latitude, longitude, and check-in hour. We create a $128 \times 128$ grid system covering the NYC area and then encode the check-in locations into grid indices. At each hour, we count the visit frequency of locations and normalize them into a 2D probability distribution, which, as a result, yields a spatiotemporal mobility matrix $M^{stm}_{NYC}$ for the whole NYC area. For trajectory reconstruction, we assign 200 users (i.e., CATS users) to meet user diversity in the raw dataset. For each CATS user, we randomly sample an anchor location (i.e., the check-in density center of a CATS user, or a CATS-Den) based on two criteria: 1) the location has check-in records in every hour during the nighttime (8 PM to 7 AM next day); and 2) locations with more check-ins have a higher probability to be sampled. After determining CATS-Dens, we randomly sample check-in locations (sampling following distance decay centered to CATS-Dens, without replacement) from $M^{stm}_{NYC}$ at each hour, disaggregating $M^{stm}_{NYC}$ into spatiotemporal mobility matrices $M^{stm}_u$ for all CATS users.

Next, we use a pre-trained CATS model to reconstruct trajectory data conditional on $M^{stm}_u$ for each CATS user, yielding a reconstructed trajectory dataset. We further analyze and compare collective-level mobility measure and OD transportation demands between the reconstructed trajectory dataset and the raw check-in dataset. The raw check-in dataset and reconstructed trajectory dataset have a random location entropy of $0.960$ and $1.323$, respectively, and their average $SSIM=0.665$. The results show that the reconstructed dataset shares high similarity with the raw dataset both in spatiotemporal distribution and movement pattern, preserving essential characteristics that supports basic mobility data utility.

\section{Discussions}

In this part, we first discuss factors that might affect the performance of our CATS framework. Then, we discuss the privacy guarantee of our method as well as other common trajectory privacy protection methods. 
In addition, we present some thoughts on the implications of privacy-preserving data publication to Humanistic GeoAI. Finally, we summarize some limitations of our current work.

\subsection{Factors Affecting Framework Performance}

We first investigate factors that may affect the privacy preservation performance. There are two factors directly related to the privacy preservation of CATS: spatiotemporal resolution and spatiotemporal mobility matrix. In our experiments, we encode the trajectory points into a $128 \times 128$ squared grid system, which is an geographic aggregation process. Each grid has an area of around 500 $m^2$. Switching to a coarser spatial resolution may help protect location privacy of users while at the cost of sacrificing data precision and utility. In an extreme case, if we encode all trajectory points into a $1 \times 1$ grid system (all the points shares the same coordinates, i.e., the center of study area), location privacy will be greatly protected as every point is indistinguishable from the others, while, as a result, data characteristics and utility will be greatly compromised. Similarly, a coarser time resolution may enhance the privacy protection of users' trajectory data, but this comes at the expense of reduced temporal data precision and utility. Another important factor to privacy is the generation and use of the spatiotemporal mobility matrix since it carries essential spatiotemporal mobility patterns that may reveal user identity and home location. In CATS, we apply the constrained clustering on users' centroids of trajectory points to get user clusters, and each user cluster has at least $K$ users (K=5). For each user cluster, we replace each user's spatiotemporal mobility matrix with the averaged spatiotemporal mobility matrix extracted from all the users in each cluster. Such that the mobility patterns of each user are mixed with at least 4 other users, leading to K-anonymity. We further verify the privacy preservation effectiveness of this averaging method by comparing it to the results directly generated from users' raw spatiotemporal mobility matrices without averaging (NoAVG-CATS). The results show that NoAVG-CATS shares much more similar TUL scores (Macro Precision: 0.233, Macro Recall: 0.139, Macro F1 scores: 0.155, Top-1 Accuracy: 0.491, and Top-5: Accuracy 0.858) and HLC scores (Mean and Median Centroid Shift: 0.007 and 0.000, Mean and Median Medoid Shift: 0.007 and 0.000, Mean and Median Number of Home Clusters: 2.000 and 2.000) with raw dataset than CATS, implying that it cannot well protect user identity or home location as it only tries to copy raw spatiotemporal mobility patterns. This proves the privacy protection effectiveness of the KAMA component in CATS.



We then investigate factors affecting characteristics and utility preservation performance. There are three factors related to the characteristics and utility preservation effectiveness of CATS: spatiotemporal resolution, spatiotemporal mobility matrix, and trajectory topology reconstruction. Spatiotemporal resolution plays a critical role in determining the scale of human mobility that a system can capture and the level of detail it can convey. For instance, when estimating origin-destination demand in the research area, a lower spatial resolution may be too coarse to accurately discern OD flow patterns, whereas a higher spatial resolution may be too fine and sparse to effectively capture the OD flow patterns due to substantial water coverage, which accounts for approximately 22\% of the research area. Similarly, a lower temporal resolution may be too coarse to capture accurate human activity patterns, whereas a higher temporal resolution may produce too much redundant information~\citep{andersen2017sampling}. Overall, our method is generic to various spatiotemporal resolutions since we can discretize any sets of trajectories from a continuous three-dimentional space into a spatiotemporal mobility matrix with arbitrary spatiotemporal resolution. Spatiotemporal mobility matrix is naturally important to spatiotemporal characteristics and utility since it represents the distribution of aggregated mobility patterns. In CATS, we preserve such a distribution by using conditional sampling. Here we further verify this by comparing the results generated from spatiotemporal mobility matrices (denoted as $\mathcal{X}_{cats}$) and the results from two-dimensional uniform distribution (denoted as $\mathcal{X}_{rd}$) where each location has the same probability to be sampled at each time steps, and we investigate the mobility measures and OD transportation demands. The results show that, compared with $\mathcal{X}_{cats}$, the random distribution result $\mathcal{X}_{rd}$ fails to preserve collective mobility measures and movement patterns (Location Random Entropy: 6.625, average SSIM with $\mathcal{X}_{raw}$: 0.026), showing the characteristics and utility preservation effectiveness of the design of spatiotemporal mobility matrix and conditional sampling in CATS. Trajectory topology reconstruction is also important to spatiotemporal characteristics and utility as it helps preserve movement patterns and geometric properties of trajectories. In CATS, we reconstruct trajectory topology by running recurrent bipartite graph matching between the trajectory point sets in adjacent hours and minimizing the overall matching cost. The optimization of this process is achieved by the adversarial training design and is also a key contribution in CATS. We verify this design by comparing the results from recurrent bipartite graph matching (denoted as $\mathcal{X}_{cats}$) to the results from random matching while using the same trajectory k-anonymization strategy (denoted as $\mathcal{X}_{tka}$) and investigate mobility measures and OD transportation demands. For mobility measures, since changing the matching relationship does not affect the spatial distribution of trajectory points, we only investigate three individual geometric properties here: jump length, location switch, and azimuth-based tortuosity (note that the radius of gyration does not depend on the matching between trajectory points). The results show that, compared with $\mathcal{X}_{cats}$, the random matching result $\mathcal{X}_{tka}$ fails to preserve both individual-level geometric properties and movement patterns (Jump Length: 1.961 km, Location Switch: 16.286, and Azimuth-Based Tortuosity: 22.228, average SSIM with $\mathcal{X}_{raw}$: 0.762) without trajectory topology reconstruction, showing the individual-level characteristics and utility preservation effectiveness of trajectory topology reconstruction design in CATS.

\subsection{Privacy Guarantee}

As the privacy protection effectiveness of CATS is examined in experiments and discussed above, we now discuss what exactly the privacy guarantee each method provides. We adopt the concept of Geo-Indistinguishability \citep{andres2013geo}, a location privacy guarantee extended from differential privacy. A privacy protection method $K(x)$ satisfies $\epsilon$-geo-indistinguishability if and only if for all points $x$, $x' \in \mathcal{X}$:

\begin{equation}
d_{\mathcal{P}}(K(x), K(x')) \leq \epsilon d(x, x')
\end{equation}

where $d(\cdot)$ denote a distance function between two points, e.g., Euclidean distance; similarly, $d_{\mathcal{P}}(\cdot)$ denote a distance function between two distributions; and $\epsilon$ is a scaling factor (extended from the privacy budget concept in differential privacy). This means, for $x$, $x' \in \mathcal{X}$ s.t. $d(x, x')\leq r$, $d_{\mathcal{P}}(K(x), K(x'))$ should be at most $l=\epsilon r$. where $r$ is the radius centered on the protected location, and $l$ is the privacy level for $r$. If a method $K(x)$ satisfies $\epsilon$-geo-indistinguishability, we also say it enjoys $l$-privacy within $r$.

From the perspective of single-location privacy protection, a Random Perturbation method $\mathcal{U}(x)$ given a perturbation threshold $r$, the noise added to the location follows a uniform distribution $X \sim \mathcal{U}(-r,r)$, which satisfies $\epsilon$-geo-indistinguishability. A Gaussian Geomasking method $\mathcal{N}(x)$ following a Gaussian distribution $X \sim \mathcal{N}(\mu,\,\sigma^{2})$, as long as controlled, could also satisfy $\epsilon$-geo-indistinguishability. Note that Gaussian distribution without further control only satisfies $(\epsilon, \delta)$-differential privacy, where $\delta$ denotes the probability of privacy leak. Location Differential Privacy and Trajectory Differential Privacy add controlled planar Laplacian noise to the location, which also satisfies $\epsilon$-geo-indistinguishability. For our method, instead of adding noise to the location, we conditionally sample locations from an averaged spatiotemporal mobility matrix from at least k=5 users, which provides K-anonymity (K=5) and satisfies $\epsilon$-geo-indistinguishability. In other words, since it conditionally re-samples locations from a bounded data distribution, a maximum sampling radius $r_s$ can be defined, and thus it satisfies $\epsilon$-geo-indistinguishability.

From the perspective of trajectory privacy protection, as \cite{andres2013geo} pointed out, applying privacy protection methods that obfuscate single locations to a group of locations (e.g., a trajectory or a dataset) suffers from a privacy downgrade issue and does not provide $\epsilon$-geo-indistinguishability guarantee. The main reason is that locations might be correlated. In an extreme case where all trajectory points share the same location, the added noise could be easily canceled out, revealing the actual location. Thus, applying random perturbation, gaussian geomasking, and location differential privacy on trajectory privacy protection only guarantees $n\epsilon$-geo-indistinguishability, a level of privacy that scales linearly with $n$, i.e., the data size. Trajectory differential privacy treats a trajectory as a basic unit and adds same noise to each point in it, which also suffers from the scalability issue and only guarantees $n\epsilon$-geo-indistinguishability. For our method, we guarantees K-anonymity which does not scale with data size, which is a stronger privacy guarantee and still satisfies $\epsilon$-geo-indistinguishability. Furthermore, the trajectory data by our method are randomly sampled from a K-anonymized spatiotemporal distribution, which reveals only information about the underlying data distribution. From this perspective, our method is a distributional-level privacy protection method which is promising for a stronger guarantee: distributional privacy \citep{blum2013learning}.















\subsection{Implications to Humanistic GeoAI}

The recent GeoAI advances have enjoyed the dividends of the booming era of location big data \citep{janowicz2019geoai}. With machine learning and deep learning technologies as its core driver, GeoAI keeps making our data-driven world more senseable and predictable. However, the use of location data is a double-edged sword, and improper use of location data may result in violations of people's geoprivacy and create severe societal, legal, ethical, or security issues \citep{kessler2018geoprivacy,sieber2016geoai}. Responding to the call of Humanistic AI, which values the humanistic perspectives of GIS \citep{zhao2021humanistic}, we advocate for Humanistic GeoAI, meaning that GeoAI is designed to be a loyal friend to humans instead of a threat. We believe that geoprivacy is a vital part of this design and needs to be taken care of with respect. In Humanistic GeoAI, Geoprivacy and GeoAI complement and benefit each other because, on the one hand, a properly designed and widely recognized geoprivacy protection mechanism guarantees that GeoAI has access to privacy-preserved location data without barriers and can utilize them for modeling, prediction, and knowledge discovery tasks without worrying about privacy issues (i.e., Geoprivacy for GeoAI). This not only allows GeoAI to support enriched location-based services and research in normal times \citep{hu2019geoai,rao2021privacy,wang2022unsupervised}, but also help save lives in time-sensitive humanitarian or emergency events such as monitoring natural disasters \citep{han2019cities} and tracking geospatial spread of diseases \citep{hou2021intracounty, huang2022uncertainties}. On the other hand, a mature and continuously iterating GeoAI can also keep bringing new insights into geoprivacy protection from the perspective of artificial intelligence (i.e., GeoAI for Geoprivacy). In our work, we propose a privacy-preserving trajectory data publication framework empowered by deep learning approaches, and we identify our work as an example of GeoAI for Geoprivacy and a contribution to the Humanistic GeoAI, which requires the consideration of human privacy rights, social responsibilities, and data ethics in designing and using location-enabled AI systems \citep{janowicz2022geoai}.

\subsection{Limitations}

While showing its good performance in all the above-mentioned comprehensive experiments, our current work still has several limitations. First, CATS generates synthetic trajectory data based on spatiotemporal mobility matrix $\mathcal{M}^{stm}$. Although this is coarse-level aggregated human mobility data, acquiring such data and ensuring its quality still requires a certain amount of work and faces uncertainty. In our second downstream utility test, we demonstrate how to acquire $\mathcal{M}^{stm}$ from sparse check-in data when there are no existing trajectory data, but data representativeness regarding users' socioeconomic background is hard to guarantee. Second, our method is a deep-learning-based framework that requires pretraining and considerable time-intensive computing resources for the development and the applications; a scalable cyberinfrastructure for GeoAI might help speed up the process. Third, the clustering process of users is currently based on centroids of user trajectories, and centroids might not always be the best characteristic to represent the active movement zones. Other user clustering strategies will be explored in our future work. Finally, we did not involve semantic attributes (e.g., place category) due to the lack of such attributes in our experimental dataset. Apart from the spatiotemporal attributes, semantic attributes play an important role in human mobility studies, especially in modeling the transportation demands and location preference. We plan to integrate semantic attributes in our future work.

\section{Conclusion and Future Work}

In this work, we present a novel deep-learning-based GeoAI framework for privacy-preserving individual-level trajectory data generation and publication, namely Conditional Adversarial Trajectory Synthesis (CATS). CATS conducts conditional adversarial training on aggregated human mobility patterns derived from large-scale individual-level GPS trajectory data to learn how to reconstruct synthetic individual trajectory data that preserve sufficient spatoiotemporal characteristics (e.g., mobility measures and OD flow matrices) and downstream data utility (e.g., location recommendation and trajectory reconstruction). We also dig deep into the data distribution level to perform K-anonymity on people's underlying spatiotemporal distribution, which provides a strong distributional-level privacy guarantee, making the generated synthetic trajectory data eligible for serving as supplements and alternatives to raw data. The comprehensive experiment results on over 90k trajectories of 1097 mobile phone users show that our method outperforms several baselines in multiple trajectory privacy protection tasks (e.g., trajectory-user-linking and home location detection), which provides people with a powerful tool to protect their geoprivacy from being violated. The results also show that our method strikes a better balance on the trade-off among privacy preservation, spatiotemporal characteristics preservation, and downstream utility, which brings new insights into the literature on privacy-preserving GeoAI and explores data ethics in GIScience.

Our future work includes exploring the synthetic trajectory generation based on irregularly aggregated human mobility data (e.g., non-squared areas) and on more sparsely sampled data sources, investigating more user clustering strategies, incorporating semantic attributes into the mobility analysis framework, and improving the design and efficacy of the deep learning framework, etc. Last but not least, the incorporation of GeoAI foundation models learned from multimodal spatiotemporal contexts and big data for human movement trajectory generation in various environments~\citep{mai2023opportunities}.

\section*{Acknowledgment}
We acknowledge the funding support provided by the American Family Insurance Data Science Institute Funding Initiative at the University of Wisconsin-Madison. Any opinions, findings, and conclusions or recommendations expressed in this material are those of the author(s) and do not necessarily reflect the views of the funder(s).

\section*{Disclosure statement}
No potential conflict of interest was reported by the author(s).

\section*{Data and Codes Availability Statement}
The data and codes that support the findings of this study are available at the following link on figshare: \url{https://doi.org/10.6084/m9.figshare.20760970}. It is worth noting that due to the non-disclosure agreement with the data provider, we are not releasing the original individual-level GPS trajectory data but sharing the k-anonymized aggregated human mobility data used in our experiments.
 
\section*{Notes on contributors}
Jinmeng Rao is a research scientist at Mineral Earth Sciences. He received his PhD degree from the Department of Geography, University of Wisconsin-Madison. His research interests include GeoAI, Privacy-Preserving AI, and Location Privacy.

Song Gao is an associate professor in GIScience at the Department of Geography, University of Wisconsin-Madison. He holds a PhD in Geography at the University of California, Santa Barbara. His main research interests include place-based GIS, geospatial data science and GeoAI approaches to human mobility and social sensing.

Sijia Zhu is a Master student in Data Science at Columbia University. She received her bachelor degrees in Statistics and Economics from the University of Wisconsin-Madison.

\bibliographystyle{apalike}
\bibliography{references}

\end{document}